\documentclass[STIX1COL]{WileyNJD-v2}
\usepackage{moreverb}
\usepackage{amsmath,amsfonts}
\usepackage{amsmath}
\usepackage{caption}
\usepackage{subcaption}
\usepackage{mathtools}
\usepackage{mathrsfs}
\usepackage{natbib}

\newcommand\BibTeX{{\rmfamily B\kern-.05em \textsc{i\kern-.025em b}\kern-.08em
T\kern-.1667em\lower.7ex\hbox{E}\kern-.125emX}}

\articletype{Advanced Review}%

\received{<26> <10>, <2023>}
\revised{<day> <Month>, <year>}
\accepted{<day> <Month>, <year>}

\begin{document}

\title{Decoding Cognitive Health Using Machine Learning: A Comprehensive Evaluation for Diagnosis of Significant Memory Concern\protect}

\author[1]{M. Sajid}

\author[1]{Rahul Sharma}

\author[2,3]{Iman Beheshti}

\author[1]{M. Tanveer}

\author[4]{for the Alzheimer’s Disease Neuroimaging Initiative}

\authormark{Sajid \textsc{et al}}

\address[1]{\orgdiv{Department of Mathematics}, \orgname{Indian Institute of Technology Indore}, \orgaddress{\state{Simrol, Indore, 453552}, \country{India}}}

\address[2]{\orgdiv{Department of Human Anatomy and Cell Science}, \orgname{Rady Faculty of Health Sciences, University of Manitoba, Winnipeg}, \orgaddress{\state{MB R3E 0J9}, \country{Canada}}}

\address[3]{\orgdiv{Neuroscience Research Program}, \orgname{Kleysen Institute for Advanced Medicine, Health Sciences Centre}, \orgaddress{\state{Winnipeg, MB R3E 0J9}, \country{Canada}}}

\address[4]{This study used data from the Alzheimer's Disease Neuroimaging Initiative (ADNI) (\href{}{adni.loni.usc.edu}). The ADNI investigators were responsible for the design and implementation of the study, but they did not take part in the analysis or the writing of this publication. \emph{http://adni.loni.usc.edu/wp-content/uploads/how\_to\_apply/ADNI\\\_Acknowledgement\_List.pdf} has a thorough list of ADNI investigators.}

\corres{*M. Tanveer, Department of Mathematics, Indian Institute of Technology Indore, Simrol, Indore, 453552, India. \email{mtanveer@iiti.ac.in}}

\abstract[Abstract]{The timely identification of significant memory concern (SMC) is crucial for proactive cognitive health management, especially in an aging population. Detecting SMC early enables timely intervention and personalized care, potentially slowing cognitive disorder progression. This study presents a state-of-the-art review followed by a comprehensive evaluation of machine learning models within the randomized neural networks (RNNs) and hyperplane-based classifiers (HbCs) family to investigate SMC diagnosis thoroughly. Utilizing the Alzheimer's Disease Neuroimaging Initiative 2 (ADNI2) dataset, 111 individuals with SMC and 111 healthy older adults are analyzed based on T1W magnetic resonance imaging (MRI) scans, extracting rich features. This analysis is based on baseline structural MRI (sMRI) scans, extracting rich features from gray matter (GM), white matter (WM), Jacobian determinant (JD), and cortical thickness (CT) measurements. 
In RNNs, deep random vector functional link (dRVFL) and ensemble dRVFL (edRVFL) emerge as the best classifiers in terms of performance metrics in the identification of SMC. In HbCs, Kernelized pinball general twin support vector machine (Pin-GTSVM-K) excels in CT and WM features, whereas Linear Pin-GTSVM (Pin-GTSVM-L) and  Linear intuitionistic fuzzy TSVM (IFTSVM-L) performs well in the JD and GM features sets, respectively. This comprehensive evaluation emphasizes the critical role of feature selection and model choice in attaining an effective classifier for SMC diagnosis. The inclusion of statistical analyses further reinforces the credibility of the results, affirming the rigor of this analysis. The performance measures exhibit the suitability of this framework in aiding researchers with the automated and accurate assessment of SMC. The source codes of the algorithms and datasets used in this study are available at
\emph{https://github.com/mtanveer1/SMC}.}

\maketitle

\section{Introduction}
\label{introduction}
Memory is a fundamental cognitive function essential for daily activities and deeply intertwined with our identities. As the global population ages, there is a growing concern about age-related cognitive decline, specifically in the form of significant memory concern (SMC). SMC, or subjective cognitive decline (SCD),  is a self-perceived decline in cognitive abilities without objective evidence of impairment. It is a common complaint among older individuals and may reflect an internal recognition of functional loss that has been compensated for. Recent research suggests that SCD may be an early indicator of more severe cognitive decline, including mild cognitive impairment (MCI) or dementia \cite{rabin2017subjective}. In other words, SMC refers to subjective complaints of memory impairment that are evident to individuals or their close associates but may not meet the criteria for clinical diagnosis of MCI or dementia \cite{jessen2020characterisation}. 

As AD pathology develops in the brain over several years before the beginning of MCI, those with SMCs may experience a distinct type of memory impairment that is distinct from the normal aging process, even before the onset of MCI \cite{gauthier2006mild}. In addition, during the earliest stages of AD, increased neuronal activity possibly caused by SMCs may lead to normal performance on cognitive tests.  Evidence has demonstrated that indicators of SMC can be beneficial in forecasting cognitive deterioration in the elderly \cite{risacher2015apoe}.

The importance of early diagnosis of SMC lies in the potential to implement preventive measures, lifestyle interventions, and targeted therapies to mitigate or delay further cognitive decline. Identifying individuals with SMC allows early intervention strategies such as cognitive training, lifestyle modifications, and pharmacological interventions to promote brain health and improve the overall quality of life. Moreover, early detection allows for timely monitoring of cognitive changes and the possibility of predicting the conversion to MCI or dementia. When evaluating individuals with SMC, it is essential to consider other potential causes of memory impairments, such as normal aging processes and age-related brain changes. Several brain pathologies have been associated with memory declines, such as gray matter  (GM), cortical thinning (CT), and white matter  (WM) \cite{filley1998behavioral}. These atrophies resemble the patterns observed in SMC and can indicate underlying neurodegenerative processes \cite{morrison2023white}. It is crucial to distinguish these atrophies from SMC to ensure accurate diagnosis and appropriate intervention strategies.

Applying machine learning (ML) techniques in clinical practice for diagnosing and managing SMC has been the subject of extensive research \cite{tanveer2020machine,sharma2023deep}. ML and deep learning are widely adopted for many applications of neurodegenerative diagnosis such as Alzheimer's Disease (AD) \cite{sharma2023conv,9715258}. Another approach of ML named ensemble learning is widely adopted nowadays and has a variety of applications in health care \cite{tanveer2024ensemble}. However, the translation of ML solutions into routine clinical use remains limited. This limitation can be attributed to concerns regarding the reliability and stability of certain ML models, leading to skepticism about their dependability and trustworthiness, particularly in medical applications. Trust in human clinicians, despite their fallibility, is deeply ingrained in our society, making it challenging to establish the same level of trust in ML systems. Consequently, it is essential to view ML as a research tool that can advance the field of study rather than a complete solution.

In recent years, significant efforts have been devoted to exploring and refining various classification algorithms. Advanced versions of well-established algorithms of HbCs, such as support vector machines (SVM) \cite{cortes1995support}, have been developed, including efficient extensions such as twin support vector machine (TSVM) \cite{khemchandani2007twin}, least square SVM (LSSVM) \cite{suykens1999least} and least square TSVM (LSTSVM) \cite{kumar2009least}. SVM and its derivatives are impressive classifiers due to their propensity for discovering optimal separating hyperplanes in high-dimensional feature spaces using Kernel techniques. Pinball SVM (Pin-SVM) \cite{6604389}, Pinball General TSVM (Pin-GTSVM) \cite{tanveer2019general}, Linex SVM (Linex-SVM) \cite{8723128}, and intuitionistic fuzzy TSVM (IFTSVM) \cite{8616852} are state-of-the-art (SOTA) variants of HbCs that extend the SVM framework to deal with non-linearly separable data, boost SVM's resistance to noise and outliers. Furthermore, alternative methods to HbCs, randomized neural network (RNN) based classifiers such as random vector functional links (RVFL) \cite{pao1994learning}, extreme learning machine (ELM) \cite{huang2006extreme}, and broad learning system (BLS) \cite{7987745}, and their variants have also been extensively investigated for their potential in classification approaches. The randomization process in RNNs mitigates computational costs, enhances resilience against overfitting, and renders the model well-adapted for handling high-dimensional data commonly encountered in practical applications.

While these studies have demonstrated promising results in many fields, it is important to recognize that ML models should be considered valuable tools for research and knowledge advancement within the field. Rather than being viewed as infallible solutions, ML should be integrated with clinical expertise and human judgment to ensure the most effective and reliable outcomes for SMC diagnosis and treatment. By leveraging the strengths of both human clinicians and ML systems, we can strive for more accurate and personalized approaches to early SMC detection.

The highlights of this comprehensive evaluation are as follows: 
\begin{itemize}
\item We conduct a comprehensive literature survey encompassing state-of-the-art reviews/survey papers, as well as relevant contributing articles.
\item We extract pertinent features from structural MRI (sMRI) scans executed by the CAT12 neuroimaging toolbox.
\item We rigorously evaluated the efficacy of $16$ HbCs and $13$ RNNs-based models to classify SMC from that of HC.
\item We employ Shapley (SHAP) \cite{molnar2020interpretable}, which is a widely recognized interpretability approach in the field of ML, to reveal significant insights into the different features.
\item Several statistical tests are conducted to assess the statistical significance of the compared models in the SMC diagnosis. 
\end{itemize}

The remaining sections of the paper are structured as follows: Section \ref{review} discusses a comprehensive literature survey encompassing state-of-the-art reviews/survey papers, as well as relevant contributing articles. Dataset, 3D MRI processing, and feature extraction are covered in Section \ref{dataset}. A brief overview of the classification methods and different strategies involved has been discussed in Section \ref{models}. Section \ref{experiments} deals with the experimental setup, results obtained from the data, feature importance analysis of SMC datasets, discussion and argumentative analysis of the findings. In Section \ref{conclusion}, we draw some conclusions and future trends.
%%%%%%%%%%%%%%%%%%%%%%%%%
    \section{State-of-the-art Review}
\label{review}
As of the present date, a considerable number of review/survey papers addressing AD diagnosis through ML and DL models have been published. It is imperative to contextualize our review within the existing landscape. Table \ref{tab:published_review} compares previously published review/survey papers with ours. This comparative analysis serves a dual purpose: firstly, it acknowledges and synthesizes the existing body of knowledge, providing a comprehensive understanding of the advancements in the field; secondly, it establishes the unique contribution and distinctive focus of our review paper within this evolving domain. Additionally, 15 relevant 2023 papers have also been included, as they were not incorporated in the review papers tabulated in  Table \ref{tab:published_review}.

\subsection{Literature selection}
The review/survey papers considered in this work were found using Google Scholar and Scopus search engines spanning the years 2022 to 2024 as of January 29, 2024. We used the keywords ``Alzheimer's disease diagnosis''+``Machine learning''+``Review paper''; ``Alzheimer's disease diagnosis''+``Deep learning''+``Review paper''; ``SMC''+``Machine learning''+``Review paper'';  ``SMC''+``Deep learning''+``Review paper''  to obtain the review papers. Initially, a total of 20 survey/review papers were downloaded using the aforementioned keywords. A two-stage selection approach was used to determine which articles from the downloaded papers should be included or excluded. Firstly, we assessed the titles and abstracts of non-duplicate research to exclude any studies that were not relevant to the focus of our evaluation. Subsequently, a thorough evaluation was conducted on the whole texts of the remaining articles. Following the two-stage procedure, 5 review/survey papers are eliminated, leaving a total of 15 review/survey papers and 15 contributing articles for consideration. 
\subsection{Literature survey}
Table \ref{tab:published_review} presents some of the statistics related to the published review papers with ours. Prior to delving into the statistics, a brief discussion on the key contributions of the referenced review papers and contributing articles are discussed subsequently.

The review paper \cite{Salman} provides an overview of the key ideas pertaining to artificial intelligence(AI)'s application in AD research, with a focus on ML and DL models. \cite{MuhammedNiyas2023} provides a concise overview, summarizing the diverse ML and DL methodologies applied to longitudinal data for predicting the transition from MCI to AD. The survey \cite{Singh} aims to offer a snapshot of recent research on DL models for AD diagnosis. The goal is to aid researchers in staying updated on the latest developments and reproducing prior studies for benchmarking purposes. In the systematic review \cite{Warren2023}, the authors examine the application of resting-state fMRI and DL methods for AD diagnosis. The review outlines common deep neural networks (DNNs), preprocessing, and classification methods found in the existing literature. In \cite{Bento2022}, DL models applied to large and heterogeneous brain imaging datasets have been summarized by investigating methods such as data access, data harmonization, and domain adaptation. The study \cite{Sanjay2023} emphasizes various ensemble techniques and DNN-based classifications for AD, underscoring the significance of ensemble methods in enhancing the accuracy and robustness of AD classification models. Authors in \cite{Sharma2023a} deep dive into ML and DL methodologies for early detection of AD, incorporating diverse feature extraction schemes to harness the complementary features of data from various multi-modal neuroimaging. The paper discusses feature selection, scaling, and fusion techniques, addressing challenges in crafting an ML-based AD diagnosis system. Furthermore, a thematic analysis is presented to compare ML workflows for potential diagnostic solutions. \cite{Fathi2022} discusses various deep models, modalities, feature extraction strategies, and parameter initialization methods that were examined to determine which model or strategy could yield superior performance. In \cite{GS2023, Saleem2022}, an overview of current approaches to AD diagnosis and detection is presented with a specific emphasis on the utilization of biomarkers. The advantages and disadvantages of employing ML and DL techniques for early AD detection are thoroughly discussed in \cite{GS2023}, whereas \cite{Saleem2022} reviews the current state-of-the-art in AD diagnosis using DL. The review paper \cite{Menagadevi2024} focuses on image pre-processing, emphasizing its contribution to noise removal, illumination, and intensity correction in Magnetic Resonance (MR) images. Segmentation methods are employed to extract the region of interest for AD detection. Feature extraction, crucial as inputs for classification, is extensively considered. Furthermore, various ML and ML algorithms are analyzed for the detection of AD. The systematic review \cite{Arya2023} examines articles utilizing ML and DL for early classification of normal cognitive (NC) and AD. The study delves into the details of commonly used PET (positron emission tomography) and MRI (magnetic resonance imaging) modalities for AD identification, evaluating their performance across different classifiers. In the mini-review, \cite{Shaaban2023}, the authors explore the potential of EEG in AD diagnosis, emphasizing the integration of EEG data with ML for effective assessment. The approach involves systematic EEG data processing, signal analysis, and application of classification algorithms. The review paper \cite{sharma2023deep} discusses imaging modalities, explores early AD biomarkers through neuroimaging scans, reviews widely-used online datasets, systematically details DL algorithms for precise early AD assessment, discusses the advantages and limitations of DL-based AD models, and offers future trend insights from a critical assessment. The study \cite{Shoeibi2023} offers a thorough review of brain disease detection, focusing on merging neuroimaging modalities through DL models such as generative adversarial networks (GANs), Autoencoders (AEs), recurrent neural networks (RNNs), and convolutional neural networks (CNNs). It begins with a discussion on neuroimaging modalities and the rationale for fusion, followed by an exploration of existing review papers in the domain of neuroimaging multimodalities utilizing AI techniques.

% \usepackage{graphicx}
\begin{table}[]
\centering
\caption{Published review papers on AD diagnosis.}
\label{tab:published_review}
\resizebox{\textwidth}{!}{%
\begin{tabular}{lccccccc} \hline
\textbf{Reference} &
  \textbf{\begin{tabular}[c]{@{}c@{}}Year of \\ Publication\end{tabular}} &
  \textbf{Papers Included} &
  \textbf{\begin{tabular}[c]{@{}c@{}}Models Studied: \\ ML/ DL\end{tabular}} &
  \textbf{\begin{tabular}[c]{@{}c@{}}Reviews Existing \\ Surveys\end{tabular}} &
  \textbf{\begin{tabular}[c]{@{}c@{}}Includes Detailed \\ Future Directions\end{tabular}} &
  \textbf{\begin{tabular}[c]{@{}c@{}}Special Attention \\ on SMC\end{tabular}} &
  \textbf{\begin{tabular}[c]{@{}c@{}}Comprehensive \\ Evaluation\end{tabular}} \\ \hline
\cite{Bento2022}          & 2022 & Prior to Aug. 31, 2021 & No/ Yes & No  & Yes & No  & No  \\
\cite{Fathi2022}          & 2022 & Prior to Feb. 08, 2022 & No/Yes  & No  & Yes & No  & No  \\
\cite{Singh}              & 2022 & Jan. 2020 - Jan. 2022    & No/Yes  & No  & No  & No  & No  \\
\cite{Warren2023}         & 2022 & Not Mentioned          & No/Yes  & No  & No  & No  & No  \\
\cite{Sharma2023a}        & 2022 & Not Mentioned          & Yes/Yes & No  & No  & No  & No  \\
\cite{Saleem2022}         & 2022 & Not Mentioned          & No/Yes  & No  & Yes & No  & No  \\
\cite{Salman}             & 2022 & Not Mentioned          & Yes/Yes & No  & No  & No  & No  \\
\cite{Sanjay2023}         & 2023 & Not Mentioned          & No/Yes  & No  & Yes & No  & No  \\
\cite{Shaaban2023}        & 2023 & Not Mentioned          & Yes/No  & No  & Yes & No  & No  \\
\cite{sharma2023deep}         & 2023 & 2009 - 2022            & No/Yes  & No  & Yes & No  & No  \\
\cite{MuhammedNiyas2023} & 2023 & Jan. 2017 - Aug. 2022    & Yes/Yes & No  & Yes & No  & No  \\
\cite{Shoeibi2023}        & 2023 & 2016 - 2022            & No/Yes  & No  & Yes & No  & No  \\ 
\cite{Arya2023}           & 2023 & Not Mentioned          & Yes/Yes & No  & No  & No  & No  \\
\cite{GS2023}            & 2023 & Not Mentioned          & Yes/Yes & No  & No  & No  & No  \\
\cite{Menagadevi2024}     & 2024 & 2013 - 2023            & Yes/Yes & No  & No  & No  & No  \\\hline
Ours                      & -    & 2022 - 2024              & Yes/Yes & Yes & Yes & Yes & Yes \\ \hline
\end{tabular}%
}
\end{table}

Besides the latest $15$ relevant review paper available for AD, there are other relevant papers published recently which has been not included in the review discussed. A total of 15 most relevant papers have been discussed. In \cite{avila2023deep}, the author proposed a novel DL model for identifying AD in clinical records. The dataset is preprocessed using various rebalancing methods, with a focus on widely used techniques like Random Oversampling, NearMiss, and SMOTE + TOMEK. The neural network architecture employs a pyramidal approximation, featuring an input layer, multiple hidden layers with decreasing units, and an output layer for binary classification. Performance evaluation reveals comparable results for SMOTE + TOMEK and Random Oversampling across various metrics, with Random Oversampling exhibiting superior accuracy and AUC due to its generating more patterns. Notably, the absence of rebalancing leads to poor performance. Although Near Miss yields modest results, it is contextualized by the small dataset employed. This study contributes insights into effective rebalancing strategies for AD classification. In \cite{mujahid2023efficient}, for addressing dataset imbalance and automated feature extraction, an adaptive synthetic oversampling technique was employed, achieving dataset balance. The study utilized an ensemble approach, combining VGG16 and EfficientNet models for AD detection on both imbalanced and balanced datasets. The proposed method amalgamated predictions from these models, creating a robust ensemble model that captured intricate data patterns. Concatenating the input and output of both models enhanced the ensemble model's complexity. Specifically, an ensemble of EfficientNet-B2 and VGG-16 demonstrated superior accuracy for early-stage AD diagnosis. The experimentation involved two publicly available datasets, substantiating the effectiveness of the proposed ensemble model.

Another approach proposed in \cite{adelson2023machine}, prompted the development and validation of a machine learning algorithm (MLA) using phenotypic data from individuals aged 55–88 (n = 493) for MCI detection. Employing a gradient-boosted tree ensemble method, the MLA consistently outperformed the mini-mental state examination (MMSE) and comparison models across multiple prediction windows. Achieving AUROC $\geq$ 0.857 and NPV $\geq$ 0.800 for all windows, the MLA demonstrated superior risk assessment for progression to AD within 24–48 months, surpassing standard care metrics. This study highlights the MLA's potential for precise risk evaluation, facilitating care coordination, and optimizing healthcare resources. Previous approaches incorporating metadata for AD diagnosis often grappled with issues like gender bias and normal aging effects. Addressing these in \cite{10365189}, the Multi-template Meta-information Regularized Network (MMRN) was devised. MMRN utilizes self-supervised learning through diverse spatial transformations for template variations and incorporates weakly supervised meta-information learning to disentangle metadata from class-related representations. Evaluated on ADNI and NACC cohorts, MMRN demonstrates promise in untangling confounding effects and enhancing AD diagnosis precision. Several randomized classification models have been proposed in recent studies, for example, 
% integrating BLS with intuitionistic fuzzy theory in \cite{sajid2023intuitionistic}. This approach effectively classifies CN vs MCI, CN vs AD, and MCI vs AD. 
in \cite{ganaie2023graph}, authors proposed a class imbalance scheme-based RVFL model for AD diagnosis. The model demonstrated improved performance in handling class imbalances specifically for AD diagnosis, effectively classifying CN vs MCI, CN vs AD, and MCI vs AD subjects.

Magnetic resonance imaging (MRI)-based assessment of cerebral atrophy serves as a valuable AD marker, commonly utilized through image registration techniques. However, the accuracy of AD detection hinges on precise image registration. To address this, a novel framework, RClaNet, integrates patch-based brain image registration and classification networks \cite{10361545}. This innovative approach enhances local deformation field estimation and disease risk prediction, showcasing superior performance in AD detection compared to contemporary methods across diverse datasets (OASIS-3, AIBL, ADNI). Another study presents an innovative deep-learning methodology tailored for automatic AD diagnosis using MRI datasets \cite{altwijri2023novel}. Diagnosing AD visually from MRI images faces challenges, particularly in discerning normal aging from early AD stages. Leveraging pre-trained convolutional neural networks (CNNs), the proposed approach enhances accuracy in detecting AD severity levels, crucial for early diagnosis. Refined image processing precedes training, addressing limited datasets. Comparative analysis with established CNNs (VGG16 and ResNet50) on Kaggle AD datasets validates the method's effectiveness across different disease stages, contributing to advancements in AD classification. Similar pretrained model-based approaches have been adopted in \cite{goel2023multimodal,sharma2022fdn}, where authors adopted pretrained model for feature extraction and state-of-the-art models to perform classification.

Innovatively enhancing CNN-based AD detection models, in \cite{10332232}, authors introduce a Graph Reasoning Module (GRM). GRM integrates an Adaptive Graph Transformer (AGT) block to construct a feature-based graph, a Graph Convolutional Network (GCN) block to refine the graph representation, and a Feature Map Reconstruction (FMR) block to convert learned graph details to a feature map. This module strategically captures relationships between diverse brain regions, elevating AD diagnostic performance through its adaptive graph reasoning mechanism. Another approach introducing an innovative AD classification framework is proposed in \cite{goyal2023multilayered}. This research combines transfer-learned AlexNet and LSTM for binary and multiclass classification of MR images. Acknowledging the demand for substantial training data, the study incorporates a Generative Adversarial Network (GAN) for data augmentation, mitigating overfitting concerns. The proposed model uses the ADNI dataset to demonstrate its efficacy on 2D MR image scans. This multilayered approach showcases the integration of deep learning and GAN techniques, addressing data limitations for improved AD classification outcomes.

Another novel approach is proposed in \cite{wang2023svfr}. This study introduces a novel two-phase Slice-to-Volume Feature Representation (SVFR) framework for effective AD diagnosis. Employing a slice-level feature extractor, a combination of DNN and clustering models automatically selects informative slice images and extracts features. Additionally, a joint volume-level feature generator and classifier hierarchically aggregate these features, utilizing a spatial pyramid set pooling and fusion modules. Experimental results showcase the superior performance of SVFR, outperforming state-of-the-art methods and achieving comparable results to the best-performing approach. In \cite{hao2023hypergraph}, the authors introduce a Weighted Hypergraph Convolution Network (WHGCN) for enhanced AD detection by utilizing internal correlations across different time points and capturing high-order relationships among subjects. Employing hypergraphs constructed with the K-nearest neighbor (KNN) method for sMRI data at each time point, the hypergraphs are fused based on the temporal importance. Hypergraph convolution is then applied for feature dimensionality reduction, enabling the learning of high-order subject relationships. Experimental validation is conducted on 518 subjects from the Alzheimer's Disease Neuroimaging Initiative (ADNI) database.\\

Examining Table \ref{tab:published_review} and the above discussion on existing review/survey and the contributing papers, it's evident that many papers solely focus on DL models, lacking discussions on future directions and neglecting an evaluation of existing review papers. The introductory discourse in Section \ref{introduction} emphasizes the criticality of early-stage detection and diagnosis of SMC. The relationship between AD and SMC is a complex and significant topic in the field of clinical medicine, deserving further examination. AD, a neurodegenerative ailment marked by gradual deterioration of cognitive function, frequently manifests with first indications and symptoms, such as memory-related worries. The modest indicators, included within the realm of SMC, act as possible predictors of more serious cognitive illnesses, such as AD. From a clinical standpoint, the emergence of notable memory apprehension prompts an important inquiry regarding its correlation with the progression of AD. The similarity in clinical presentation between SMC and the initial phases of AD underscores the need to comprehend their intricate interaction. Early detection of SMC is crucial in this setting, as it may indicate the beginning of underlying pathogenic mechanisms contributing to AD. However, there is a noticeable lack of research investigations that utilize modern computational models, such as ML and DL, to address the therapeutic importance of SMC. While many review articles explain the intricacies of AD which we have discussed and tabulated earlier, there is a lack of particular studies that focus on the role of SMC using ML and DL approaches. This highlights the need to do a thorough examination of different models, hence, we are motivated to explore the emerging field of SMC classification due to the detection of a gap in existing research. We are dedicated to conducting innovative research beyond conventional methods, exploring the unexplored field of ML and DL for SMC classification. An exhaustive assessment across many models aims to not only fill the existing gap in the literature but also advance the discipline by providing groundbreaking insights into the initial phases of cognitive decline. Our complete analysis aims to address a significant gap in the field of cognitive health management by combining clinical knowledge, computational methods, and the need for proactive measures. This work seeks to fill these gaps by providing a specific focus on SMC. It involves a comprehensive assessment of SMC detection utilizing ML models from two prominent families, namely RNNs and HbCs.

While our paper seeks to review the latest research in the field comprehensively, we recognize the importance of placing our findings in the broader context of existing literature to underscore their unique contributions. To address this, we aim to enrich the discussion section by juxtaposing our results with those of prior studies, demonstrating how our approach expands upon existing research and provides insights for future investigations and clinical applications. Moreover, we will emphasize the originality of our methodology, illustrating how it fills gaps in current literature and advances knowledge in the field. This strategy will ensure that readers grasp the significance of our study within the larger research landscape.
%%%%%%%%%%%%%%%%%%%%%%%%%%%%%%%%%%%%%
\section{Data Acquisition and Preprocessing}
\label{dataset}
This section discusses the utilized SMC data in the evaluation, encompassing aspects from data acquisition to preprocessing.
\subsection{Data Acquisition}
Data from the Alzheimer's Disease Neuroimaging Initiative (ADNI) dataset (ADNI2) is used to classify 111 individuals with subjective cognitive decline (SCD) (mean age $\pm$ SD: 72.31 $\pm$ 5.49 years, 57\% women) and 111 cognitively healthy older adults (mean age $\pm$ SD: 73.36 $\pm$ 6.36 years, 53\% women) and based on their baseline structural MRI scans. There was no significant difference between the two groups in terms of age (P $=$ 0.15, t-test) or sex (P $=$ 0.56, chi-squared test).
\subsection{Data Preprocessing}
The CAT12 toolbox \emph{(http://www.neuro.uni-jena.de/cat/)}, which is an extension of the SPM12 software package, is used to process the structural MRI scans in this study. The raw MRI scans underwent several preprocessing steps, including bias correction to reduce intensity variations and segmentation into gray matter (GM), white matter (WM), and cerebrospinal fluid (CSF) images. The DARTEL normalization to the Montreal Neurological Institute (MNI) space was applied (voxel size of $1.5$ mm $\times$ $1.5$ mm $\times$ $1.5$ mm), and modulation is also employed to preserve volumetric information \cite{farokhian2017comparing}. Jacobian determinant (JD) images are also generated. Image quality and brain segmentation are visually inspected by I.B. as well as the "Check Homogeneity" feature of the CAT12 toolbox. Gaussian smoothing with a full width at half maximum (FWHM) of 4 mm is applied to all generated images, including GM, WM, and JD, to improve spatial coherence and increase the signal-to-noise ratio. Employing the Brainnetome atlas, the mean regional signals are calculated, and $273$ features across GM, WM, and JD are obtained for each subject.  The cortical thickness (CT) measurements are also extracted based on the Desikan-Killiany-Tourville (DKT) atlas, comprising $68$ features across the cortical area \cite{klein2012101}. Additionally, age and sex are the demographic features that are also included in each modality's feature set. 

\noindent \textbf{Note:} The datasets used in this study are available at
\emph{https://github.com/mtanveer1/SMC}.
%%%%%%%%%%%%%%%%%%%%%%%%%%%%
\section{Classification Models}
\label{models}
\begin{figure*}[ht!]
    \centering
    \includegraphics[width=0.7\linewidth]{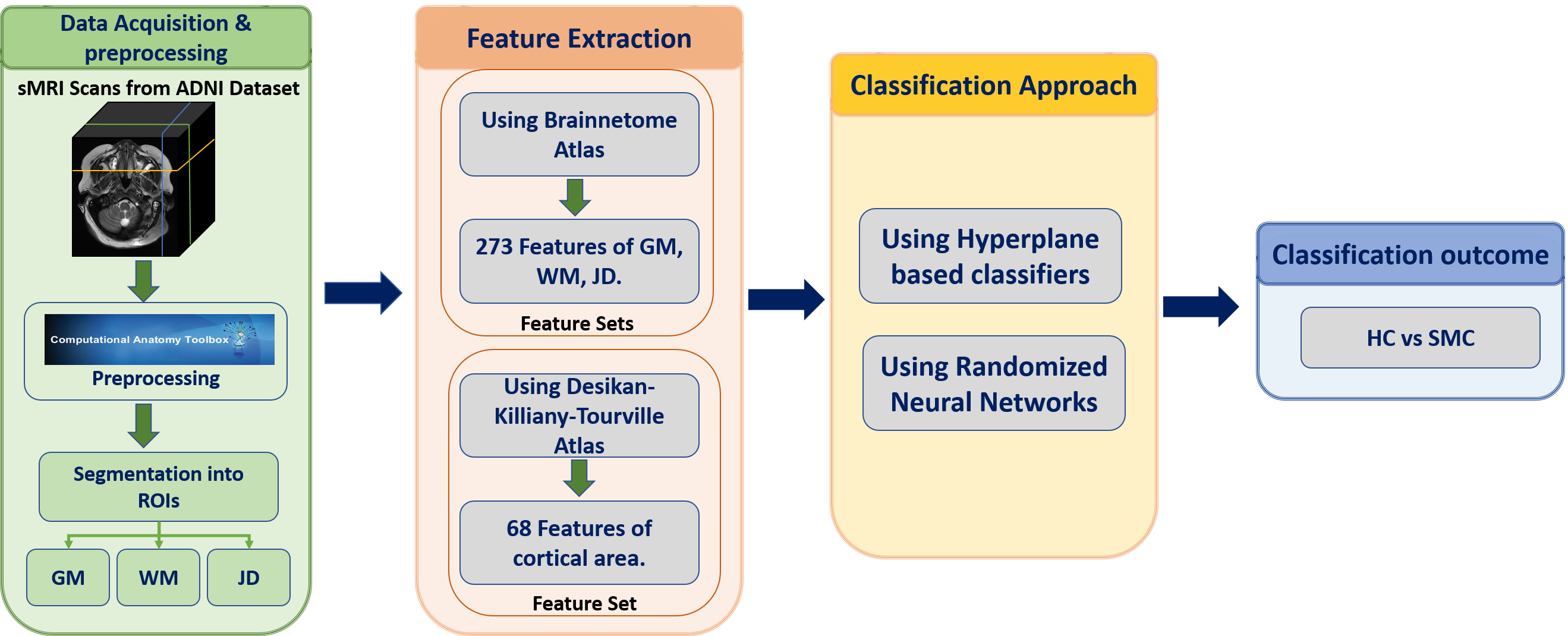}
    \caption{Flow diagram of the proposed comprehensive evaluation approach.}
    \label{fig1}
\end{figure*}

The study assesses classification algorithms for CN vs. SMC, categorizing them into RNNs and HbCs. By embracing ML classifiers from two diverse domains: (a) HbCs with strong mathematical foundation and excellent adaptability of Kernel tricks and (b) RNNs acclaimed for their rapid learning, universal approximation property \cite{pao1994learning, huang2006extreme, 7987745} and higher generalization performance provide solid backup and multiple perspectives to this comprehensive SMC assessment. This comparison highlights the distinctions between the two approaches, delving into their strengths and limitations in SMC diagnosis. A detailed description of the approach has been represented in Figure \ref{fig1} and details of each classifier, followed by the rationale behind selecting these classifiers, is elaborated below.
\subsection{Randomized Neural Networks (RNNs)}
In this assessment, we included a total of $13$ state-of-the-art (SOTA) RNN-based classifiers, each of which is succinctly elaborated upon below:
\begin{enumerate}
    \item \textbf{Random Vector Functional Link (RVFL) Neural Network:} The RVFL \cite{pao1994learning, malik2022random}  is a shallow RNN classifier in which the hidden layer's weights and biases are initialized randomly and remain constant during training. The output layer's parameters, which include both the direct link weights (the connection between the input layer and the output layer) and the connections between the hidden layer and the output layer, are determined analytically using methods such as the Pseudo-inverse or least-square technique. The inclusion of direct links in the RVFL has been found to enhance learning performance notably, serving as a form of regularization to complement the inherent randomization in the model \cite{zhang2016comprehensive}.
    \item \textbf{Extreme Learning Machine (ELM):} The ELM \cite{huang2006extreme} is a single hidden layer feedforward neural network (SLFN) that randomly chooses the input weights and analytically determines the output weights of SLFN. Distinctively, ELM differs from the RVFL model by not incorporating direct links.
    \item \textbf{Minimum Class Variance Extreme Learning Machine (MCVELM):} The MCVELM \cite{6542653}, a variant of ELM, is designed to minimize both the norm of network output layer weights and the intraclass variance of training data within the decision space. This unique approach contributes to enhancing the model's discriminative capabilities and improving its overall performance in handling complex data distributions.
    \item \textbf{Minimum Variance Extreme Learning Machine (MVELM):} The MVELM \cite{iosifidis2014minimum} classifier strategically leverages the dispersion within the training data during its optimization process. This utilization of data spread enhances the classifier's adaptability and contributes to improved performance in capturing the underlying patterns and nuances within the dataset.
    \item \textbf{Intuitionistic Fuzzy Random Vector Functional Link (IFRVFL):} The IFRVFL \cite{9715258} adopts a fuzzy weighted strategy for optimal classifier generation, assigning each sample an intuitionistic fuzzy number (IFN) based on membership and nonmembership scores. The membership score considers the sample's distance from its class centroid, while the nonmembership score considers both distances from the centroid and neighborhood information. IFRVFL reduces the negative impact of noise and outliers and enhances model generalization capability.
    \item \textbf{Total Variance Minimization
    based Random Vector Functional Link Network (Total-Var-RVFL):} The Total-Var-RVFL \cite{ganaie2020minimum} classifier optimizes output layer weights by minimizing total variance, leveraging training data dispersion in both the original feature space and the randomized feature projection space.
    \item \textbf{Intraclass Variance Minimization based Random Vector Functional Link Network (Class-Var-RVFL):} The Class-Var-RVFL \cite{ganaie2020minimum} optimizes output weights by minimizing the intraclass variance of the training data in both original and randomized feature spaces, enhancing model's generalization performance across diverse data distributions.
    \item \textbf{Graph Embedded Extreme Learning Machine:}
    The GEELM \cite{7052327} model integrates subspace learning (SL) criteria into the optimization process for calculating network output weights. The GEELM adeptly leverages both intrinsic and penalty SL criteria within the graph embedding (GE) framework. GEELM-LDA \cite{7052327} employs linear discriminant analysis graph embedding, while GEELM-LFDA \cite{7052327} employs the local Fisher discriminant analysis GE framework.
    \item \textbf{Deep Random Vector Functional Link (dRVFL):} The dRVFL \cite{shi2021random}, an expansion of RVFL focusing on representational learning, is composed of stacked hidden layers. The parameters of hidden layers are randomly generated and held constant, with only the output weights undergoing computation. This design maintains fixed internal representations of RVFL, emphasizing adaptability in learning complex non-linear patterns.
    \item \textbf{Ensemble Deep Random Vector Functional Link (edRVFL):} The edRVFL \cite{shi2021random} model is a synthesis of ensembling learning and deep learning that employs an implicit ensembling strategy. The ensemble comprises $L$ models, aligning with the number of hidden layers present in a single dRVFL model. The ensemble is trained such that the higher models (corresponding to deeper layers in the dRVFL network) employ non-linearly transformed features from the antecedent layers as well as the original features (derived from direct connections as in the standard RVFL network). Therefore, the framework simultaneously adheres to the principles of deep learning and ensemble learning. 
    \item \textbf{Broad Learning System (BLS):} The BLS \cite{7987745} provides an alternative approach to deep models. In BLS, the initial inputs are transmitted to ``mapped features'' in feature nodes, while the structure is broadly expanded in ``enhancement nodes.'' This results in a network that is quick, efficient, and flat. The weights of the output layer are computed using the Pseudo-inverse.
    \item \textbf{Neuro Fuzzy Broad Learning System (NF-BLS):} The NF-BLS \cite{8432091} represents a fuzzy adaptation of BLS, where the feature nodes in BLS are substituted with first-order Takagi–Sugeno (TS) fuzzy subsystems. Parameters in NF-BLS encompass weights linking the enhancement layer outputs to the final output layer and coefficients in the consequent part of fuzzy rules within each fuzzy subsystem. These parameters are calculated through Pseudo-inverse methods.
\end{enumerate}
%%%%%%%%%%%%%%%%%%%%%%%%%%
\subsubsection{Rationale behind behind selecting the RNN classifiers}
In RNN, some network parameters are fixed during the training period, and parameters of output layers are calculated via the closed-form solution. The selected models exhibit a diverse nature, each offering a unique set of features tailored to address specific challenges. From handling noise and outliers to capturing intricate geometric relationships and minimizing various forms of variance, these models collectively contribute to a comprehensive and nuanced evaluation of the SMC dataset. Among these, RVFL \cite{pao1994learning}, ELM \cite{huang2006extreme}, and BLS \cite{7987745} stand out as prominent representatives within the RNN family, each offering distinct advantages and unique features. Notably, standard RVFL, ELM, and BLS follow a uniform weightage scheme for each sample, leading to challenges in handling noise and outliers present in datasets. To address these challenges, we incorporate IFRVFL \cite{9715258} as a classifier in our assessment. IFRVFL excels in handling noise and outliers, providing a robust solution. RNNs' randomization process often disregards the inherent geometric relationships of data and topological properties during the computation of final output parameters. To mitigate this, we consider GEELM \cite{7052327} in our evaluation. GEELM adeptly leverages both intrinsic and penalty SL criteria within the GE framework, providing a nuanced understanding of the dataset's structural properties. Additionally, to minimize the variance of each class, we incorporate Class-Var-RVFL \cite{ganaie2020minimum} and MCVELM \cite{6542653} in our evaluation. For total variance minimization of the dataset, MVELM \cite{iosifidis2014minimum} and Total-Var-RVFL \cite{ganaie2020minimum} are selected. In addressing the ``black-box'' nature of artificial neural networks (ANNs), especially in critical applications like healthcare, where interpretability is crucial, we take neuro NF-BLS \cite{8432091}. This classifier adopts an IF-THEN approach, providing transparency to the decision-making process to some extent. Recognizing the limitations of single-layer RVFL in extracting non-linear features, we consider dRVFL \cite{shi2021random}. Enabling multiple hidden layers to collaboratively extract meaningful features from datasets, dRVFL enhances the model's capacity. In the rapidly evolving landscape of machine learning, ensemble deep learning \cite{ganaie2022ensemble} has emerged as a pivotal advancement; therefore, edRVFL \cite{shi2021random} is a unique fusion of deep learning and ensemble learning principles, contributing a distinctive dimension to our assessment.
%%%%%%%%%%%%
\subsection{Hyperplane-based Classifiers (HbCs)}
In the course of this evaluation, $16$ SOTA HbC models under examination are subjected to evaluation within both linear and non-linear Kernel spaces. The model $\mathcal{M}$ in the linear space is designated as $\mathcal{M}$-L, whereas its counterpart assessed in the Kernel space is referred to as $\mathcal{M}$-K. Gaussian Kernel is employed for the Kernel-based models. All the HbC evaluated in this study are briefly explained below in general (for linear as well as Kernel space):
\begin{enumerate}
    \item \textbf{Support Vector Machine (SVM):} The Support Vector Machine (SVM) \cite{cortes1995support} is a formidable machine learning model meticulously designed to maximize the margin between two classes in classification scenarios. Its objective is to ascertain the optimal hyperplane situated between two parallel supporting hyperplanes. The SVM's monumental success is attributed to its adept incorporation of the Kernel approach and strong mathematical and statistical foundation. The optimal hyperplane in SVM is determined by solving a convex quadratic programming problem (QPP).
    \item \textbf{Twin Support Vector Machine (TSVM):} The TSVM \cite{khemchandani2007twin}, a binary classifier, establishes two nonparallel hyperplanes through the resolution of two smaller-sized Quadratic Programming Problems (QPPs), in contrast to a single large QPP as seen in SVM. This approach makes TSVM four times more efficient than the SVM \cite{khemchandani2007twin}.
    \item \textbf{Intuitionistic Fuzzy Twin Support Vector Machine (IFTSVM):} The IFTSVM \cite{8616852}, a fuzzy adaptation of TSVM, categorizes input samples using degrees of both membership and nonmembership functions. This dual-functionality serves to mitigate the impact of noise and outliers, contributing to enhanced robustness in classification.
    \item \textbf{Least Square Support Vector Machine (LSSVM):} The LSSVM \cite{suykens1999least} addresses classification problems through the inclusion of equality constraints in the optimization problem, formulating it in the least squares sense. Consequently, the solution is obtained directly by solving a system of linear equations, bypassing the need to solve QPP/QPPs typical in SVM approaches.
    \item \textbf{Least Square Twin Support Vector Machine (LSTSVM):} The LSTSVM \cite{kumar2009least}, using a similar approach to LSSVM, incorporates equality constraints in the optimization problem. As a result, LSTSVM only necessitates solving two systems of linear equations in contrast to TSVM, which requires solving QPPs and two systems of linear equations.
    \item \textbf{Linex Support Vector Machine (Linex-SVM):} The Linex-SVM \cite{8723128} incorporates the asymmetric Linex (linear exponential) loss function, which imposes a more substantial penalty on points situated between two central planes while administering a milder penalty to points located outside the corresponding central planes. The Linex loss in SVM contributes to the model's enhanced generalization performance.
    \item \textbf{Pinball Support Vector Machine (Pin-SVM):} The Pin-SVM \cite{6604389} integrates the Pinball loss function, factoring in quantile distance to impose penalties on input samples. This design, characterized by its resilience to noise and heightened stability during re-sampling, underscores the model's robustness and diminished sensitivity.
    \item \textbf{Pinball General Twin Support Vector Machine (Pin-GTSVM):} The Pin-GTSVM \cite{tanveer2019general} is a pinball loss variant of the TSVM classifier and is renowned for its robustness against noise and stability during resampling. 
\end{enumerate} 
\subsubsection{Rationale behind behind selecting the HbCs}
In HbCs, the classification process involves segregating data points belonging to distinct classes by utilizing hyperplanes. We choose a set of $16$ classifiers from the realm of HbCs. The SVM \cite{cortes1995support} is the first HbCs in which binary datasets are classified using the concept of optimal hyperplane. Solid mathematical and statistical underpinnings SVM. SVM solves one large quadratic programming problem (QPP), resulting in escalated computational complexity, which renders it less suitable for large-scale datasets. To alleviate the time complexity of SVM, TSVM is proposed \cite{khemchandani2007twin}. TSVM solves two smaller-sized QPPs instead of a single large QPP, making TSVM four times faster than the standard SVM, and then we consider TSVM as well in our assessment. Standard SVM and TSVM use a uniform weightage strategy for each sample, which creates difficulties when dealing with noise and outliers in datasets. As a classifier, we consider IFTSVM \cite{8616852} to deal with these issues. The computational cost of SVM and TSVM is increased since they both tackle QPP in their optimization problem to obtain the ideal hyperplane(s). As a solution, we add LSSVM \cite{suykens1999least} and LSTSVM \cite{kumar2009least} into our analysis. Instead of solving QPPs, as is required by SVM and TSVM, LSSVM solves system(s) of linear equations. Misclassified samples are punished in the standard SVM, TSVM, and LSVM by means of the hinge loss function. Hinge loss makes the models less robust to noise and less stable over resampling \cite{6604389}. HbCs based on the Linex loss and the pinball loss function have been offered as solutions. In this comprehensive evaluation, we take into account Linex-SVM \cite{8723128}, Pin-SVM \cite{6604389} and Pin-GTSVM \cite{tanveer2019general} in our comprehensive evaluation. The SVM is known for its adaptability of the Kernel technique in its structure. To keep this thing in mind, we consider Kernalized variants as well of all the above-mentioned HbCs. Kernelized variants help to extract the non-linear, non-separable datasets for classification.\\

\section{Experimental Evaluation and Discussion}
\label{experiments}
In this section, we provide an intricate overview of the experimental setup, conduct a thorough comparison among HbCs and RNNs on SMC datasets within each modality, \textit{i.e.,}, CT, GM, JD, WM, and fusion of all Features of each modality (referred to as All Features), and perform a comprehensive statistical analysis. Our analysis delves into the examination of crucial features within each modality,  subsequently followed by a detailed discussion of the obtained results. The source codes of the algorithms and datasets used in this study are available at
\emph{https://github.com/mtanveer1/SMC}.

\subsection{Experimental Setup}
In this subsection, we provide the implementation details, experimental setup, hyperparameter configurations, and performance metrics employed for the thorough assessment of the models.
% \usepackage{graphicx}
\begin{table}[ht!]
\centering
\caption{Parameters' description and range for RNN-based classifiers.}
\label{tab:Randomized_Parameters_Range}
\resizebox{0.84\linewidth}{!}{
\begin{tabular}{lll}
\hline
\multicolumn{1}{|l|}{RNN} & \multicolumn{1}{c|}{Parameters' description} & \multicolumn{1}{c|}{Parameters' Range} \\ \hline
\multicolumn{1}{|l|}{RVFL} &
  \multicolumn{1}{l|}{\begin{tabular}[c]{@{}l@{}}$C$: Regularization parameter\\ $N$: Number of hidden nodes\\ $Act$: Activation function\end{tabular}} &
  \multicolumn{1}{l|}{\begin{tabular}[c]{@{}l@{}}$C=[10^{-8},10^{-6},10^{-4},10^{-2},1,10^2,10^4,10^6,10^8]$\\ $N=3:20:503$\\ $Act=1:1:9$\end{tabular}} \\ \hline
\multicolumn{1}{|l|}{ELM} &
  \multicolumn{1}{l|}{\begin{tabular}[c]{@{}l@{}}$C$: Regularization parameter\\ $N$: Number of hidden nodes\\ $Act$: Activation function\end{tabular}} &
  \multicolumn{1}{l|}{\begin{tabular}[c]{@{}l@{}} $C=[10^{-8},10^{-6},10^{-4},10^{-2},1,10^2,10^4,10^6,10^8]$\\ $N=3:20:503$\\ $Act=1:1:9$\end{tabular}} \\ \hline
\multicolumn{1}{|l|}{MCVELM}     & \multicolumn{1}{l|}{\begin{tabular}[c]{@{}l@{}}$C$: Regularization parameter\\ $\lambda$: Variance regularization parameter\\ $N$: Number of hidden nodes\\ $Act$: Activation function\end{tabular}}     & \multicolumn{1}{l|}{\begin{tabular}[c]{@{}l@{}}$C=[10^{-8},10^{-6},10^{-4},10^{-2},1,10^2,10^4,10^6,10^8]$\\ $\lambda=[10^{-8},10^{-6},10^{-4},10^{-2},1,10^2,10^4,10^6,10^8]$\\ $N=3:20:503$\\ $Act=1:1:9$\end{tabular}}\\ \hline
\multicolumn{1}{|l|}{MVELM}    & \multicolumn{1}{l|}{\begin{tabular}[c]{@{}l@{}}$C$: Regularization parameter\\ $\lambda$: Variance regularization parameter\\ $N$: Number of hidden nodes\\ $Act$: Activation function\end{tabular}}       & \multicolumn{1}{l|}{\begin{tabular}[c]{@{}l@{}}$C=[10^{-8},10^{-6},10^{-4},10^{-2},1,10^2,10^4,10^6,10^8]$\\ $\lambda=[10^{-8},10^{-6},10^{-4},10^{-2},1,10^2,10^4,10^6,10^8]$\\ $N=3:20:503$\\ $Act=1:1:9$\end{tabular}}\\ \hline 
\multicolumn{1}{|l|}{IFRVFL}     & \multicolumn{1}{l|}{\begin{tabular}[c]{@{}l@{}}$C$: Regularization parameter\\ $\mu$: Intuitionistic fuzzy Kernel parameter\\ $N$: Number of hidden nodes\\ $Act$: Activation function\end{tabular}}      & \multicolumn{1}{l|}{\begin{tabular}[c]{@{}l@{}}$C=[10^{-8},10^{-6},10^{-4},10^{-2},1,10^2,10^4,10^6,10^8]$\\$\mu=[10^{-5},10^{-4},10^{-3},10^{-2},10^{-1},1,10^1,10^2,10^3,10^4,10^5]$\\ $N=3:20:503$\\ $Act=1:1:9$\end{tabular}}\\ \hline 
\multicolumn{1}{|l|}{Class-Var-RVFL}   & \multicolumn{1}{l|}{\begin{tabular}[c]{@{}l@{}}$C$: Regularization parameter\\ $\lambda$: Variance regularization parameter\\ $N$: Number of hidden nodes\\ $Act$: Activation function\end{tabular}}        & \multicolumn{1}{l|}{\begin{tabular}[c]{@{}l@{}}$C=[10^{-8},10^{-6},10^{-4},10^{-2},1,10^2,10^4,10^6,10^8]$\\ $\lambda=[10^{-8},10^{-6},10^{-4},10^{-2},1,10^2,10^4,10^6,10^8]$\\ $N=3:20:503$\\ $Act=1:1:9$\end{tabular}}\\ \hline    
\multicolumn{1}{|l|}{Total-Var-RVFL}     & \multicolumn{1}{l|}{\begin{tabular}[c]{@{}l@{}}$C$: Regularization parameter\\ $\lambda$: Variance regularization parameter\\ $N$: Number of hidden nodes\\ $Act$: Activation function\end{tabular}}      & \multicolumn{1}{l|}{\begin{tabular}[c]{@{}l@{}}$C=[10^{-8},10^{-6},10^{-4},10^{-2},1,10^2,10^4,10^6,10^8]$\\ $\lambda=[10^{-8},10^{-6},10^{-4},10^{-2},1,10^2,10^4,10^6,10^8]$\\ $N=3:20:503$\\ $Act=1:1:9$\end{tabular}}\\ \hline
\multicolumn{1}{|l|}{GEELM-LDA}   & \multicolumn{1}{l|}{\begin{tabular}[c]{@{}l@{}}$C$: Regularization parameter\\ $\lambda$: Graph regularization parameter\\ $N$: Number of hidden nodes\\ $Act$: Activation function\end{tabular}}        & \multicolumn{1}{l|}{\begin{tabular}[c]{@{}l@{}}$C=[10^{-8},10^{-6},10^{-4},10^{-2},1,10^2,10^4,10^6,10^8]$\\ $\lambda=[10^{-6},10^{-4},10^{-2},1,10^2,10^4,10^6]$\\ $N=3:20:503$\\ $Act=1:1:9$\end{tabular}}\\ \hline
\multicolumn{1}{|l|}{GEELM-LFDA}   & \multicolumn{1}{l|}{\begin{tabular}[c]{@{}l@{}}$C$: Regularization parameter\\ $\lambda$: Graph regularization parameter\\ $N$: Number of hidden nodes\\ $Act$: Activation function\end{tabular}}        & \multicolumn{1}{l|}{\begin{tabular}[c]{@{}l@{}}$C=[10^{-8},10^{-6},10^{-4},10^{-2},1,10^2,10^4,10^6,10^8]$\\ $\lambda=[10^{-6},10^{-4},10^{-2},1,10^2,10^4,10^6]$\\ $N=3:20:503$\\ $Act=1:1:9$\end{tabular}}\\ \hline
\multicolumn{1}{|l|}{dRVFL}     & \multicolumn{1}{l|}{\begin{tabular}[c]{@{}l@{}}$C_1$: Regularization parameter for the first stage tuning\\ $N_1$: Number of hidden nodes for the first stage tuning\\ $L_1$: Number of hidden layers for the first stage tuning\\ $Act_1$: Activation function for the first stage tuning \\ \hline 
$C_2$: Regularization parameter for the second stage tuning\\
$N_2$: Number of hidden nodes for the second stage tuning\\
$L_2$: Number of hidden layers for the second stage tuning \\
$Act_2$: Activation function for the second stage tuning\end{tabular}}      & \multicolumn{1}{l|}{\begin{tabular}[c]{@{}l@{}}$C_1=[10^{-8},10^{-6},10^{-4},10^{-2},1,10^2,10^4,10^6,10^8]$\\ $N_1=[256,512,1024]$\\ $L_1=2$\\ $Act_1=7$ \\ \hline
$C_2=C_1^{*}\times[0.9,0.925,0.95,0.975,1,1.025,1.05,1.075,1.1]$\\
$N_2=N_1^{*}\times[0.9,0.925,0.95,0.975,1,1.025,1.05,1.075,1.1]$\\
$L_2=1:1:10$ \\
$Act_2=1:1:9$\end{tabular}}\\ \hline
\multicolumn{1}{|l|}{edRVFL}     & \multicolumn{1}{l|}{\begin{tabular}[c]{@{}l@{}}$C_1$: Regularization parameter for the first stage tuning\\ $N_1$: Number of hidden nodes for the first stage tuning\\ $L_1$: Number of hidden layers for the first stage tuning\\ $Act_1$: Activation function for the first stage tuning \\ \hline 
$C_2$: Regularization parameter for the second stage tuning\\
$N_2$: Number of hidden nodes for the second stage tuning\\
$L_2$: Number of hidden layers for the second stage tuning \\
$Act_2$: Activation function for the second stage tuning\end{tabular}}      & \multicolumn{1}{l|}{\begin{tabular}[c]{@{}l@{}}$C_1=[10^{-8},10^{-6},10^{-4},10^{-2},1,10^2,10^4,10^6,10^8]$\\ $N_1=[256,512,1024]$\\ $L_1=2$\\ $Act_1=7$ \\ \hline
$C_2=C_1^{*}\times[0.9,0.925,0.95,0.975,1,1.025,1.05,1.075,1.1]$\\
$N_2=N_1^{*}\times[0.9,0.925,0.95,0.975,1,1.025,1.05,1.075,1.1]$\\
$L_2=1:1:10$\\
$Act_2=1:1:9$\end{tabular}}\\ \hline
\multicolumn{1}{|l|}{BLS}    & \multicolumn{1}{l|}{\begin{tabular}[c]{@{}l@{}}$C$: Regularization parameter\\$ N_{Feat-G}$: Number of feature groups\\$N_{Feat-N}$: Number of feature nodes in each groups\\ $N_{E-G}$: Number of enhancement groups\\$N_{E-N}$: Number of enhancement nodes in each groups\\ $Act$: Activation function\end{tabular}}       & \multicolumn{1}{l|}{\begin{tabular}[c]{@{}l@{}}$C=[10^{-8},10^{-6},10^{-4},10^{-2},1,10^2,10^4,10^6,10^8]$\\$ N_{Feat-G}=5:5:50$\\$N_{Feat-N}=1:2:21$\\ $N_{E-G}=5:10:105$\\$N_{E-N}=1$\\ $Act=1:1:9$\end{tabular}}\\ \hline
\multicolumn{1}{|l|}{NF-BLS}    & \multicolumn{1}{l|}{\begin{tabular}[c]{@{}l@{}}$C$: Regularization parameter\\$ N_{Feat-G}$: Number of fuzzy groups\\$N_{Feat-N}$: Number of fuzzy nodes in each groups\\ $N_{E-G}$: Number of enhancement groups\\$N_{E-N}$: Number of enhancement nodes in each groups\\ $Act$: Activation function\end{tabular}}       & \multicolumn{1}{l|}{\begin{tabular}[c]{@{}l@{}}$C=[10^{-8},10^{-6},10^{-4},10^{-2},1,10^2,10^4,10^6,10^8]$\\$N_{Fuzz-G}=5:5:50$\\$N_{Fuzz-N}=1:2:21$\\ $N_{E-G}=5:10:105$\\$N_{E-N}=1$\\ $Act=1:1:9$\end{tabular}}\\ \hline
\end{tabular}
}
\end{table}
% \usepackage{graphicx}
\begin{table}[ht!]
\centering
\caption{Parameters' description and range for HbCs.}
\label{tab:hyperplane_Parameters_Range}
\resizebox{15cm}{!}{
\small
\begin{tabular}{lll}
\hline
\multicolumn{1}{|l|}{HbC} & \multicolumn{1}{c|}{Parameters' description} & \multicolumn{1}{c|}{Parameters' Range}\\ \hline
\multicolumn{1}{|l|}{SVM-L} &
  \multicolumn{1}{l|}{\begin{tabular}[c]{@{}l@{}}$C$: Regularization parameter\end{tabular}} &
  \multicolumn{1}{l|}{\begin{tabular}[c]{@{}l@{}}$C=[2^{-5},2^{-3},2^{-1},2^{1},2^3,2^5]$ \end{tabular}} \\ \hline
\multicolumn{1}{|l|}{SVM-K} &
  \multicolumn{1}{l|}{\begin{tabular}[c]{@{}l@{}}$C$: Regularization parameter\\ $\sigma$: Kernel parameter\end{tabular}} &
  \multicolumn{1}{l|}{\begin{tabular}[c]{@{}l@{}}$C=[2^{-5},2^{-3},2^{-1},2^{1},2^3,2^5]$ \\ $\sigma=[2^{-10},2^{-9},\cdots,2^{10}]$ \end{tabular}} \\ \hline
  \multicolumn{1}{|l|}{TSVM-L} &
  \multicolumn{1}{l|}{\begin{tabular}[c]{@{}l@{}}$C_1$: Regularization parameter for the positive class\\ $C_2$: Regularization parameter for the negative class \end{tabular}} &
  \multicolumn{1}{l|}{\begin{tabular}[c]{@{}l@{}}$C_1=[2^{-5},2^{-3},2^{-1},2^{1},2^3,2^5]$  \\ $C_2=[2^{-5},2^{-3},2^{-1},2^{1},2^3,2^5]$ \end{tabular}} \\ \hline
  \multicolumn{1}{|l|}{TSVM-K} &
  \multicolumn{1}{l|}{\begin{tabular}[c]{@{}l@{}}$C_1$: Regularization parameter for the positive class\\ $C_2$: Regularization parameter for the negative class\\ $\sigma$: Kernel parameter  \end{tabular}} &
  \multicolumn{1}{l|}{\begin{tabular}[c]{@{}l@{}}$C_1=[2^{-5},2^{-3},2^{-1},2^{1},2^3,2^5]$  \\ $C_2=[2^{-5},2^{-3},2^{-1},2^{1},2^3,2^5]$ \\ $\sigma=[2^{-10},2^{-9},\cdots,2^{10}]$ \end{tabular}} \\ \hline
   \multicolumn{1}{|l|}{IFTSVM-L} &
  \multicolumn{1}{l|}{\begin{tabular}[c]{@{}l@{}}$C_1$: Regularization parameter for the positive class\\ $C_2$: Regularization parameter for the negative class \\ $\mu$: Intuitionistic fuzzy Kernel parameter \end{tabular}} &
  \multicolumn{1}{l|}{\begin{tabular}[c]{@{}l@{}}$C_1=[2^{-5},2^{-3},2^{-1},2^{1},2^3,2^5]$  \\ $C_2=[2^{-5},2^{-3},2^{-1},2^{1},2^3,2^5]$ \\ $\mu=[2^{-10},2^{-9},\cdots,2^{10}]$ \end{tabular}} \\ \hline
   \multicolumn{1}{|l|}{IFTSVM-K} &
  \multicolumn{1}{l|}{\begin{tabular}[c]{@{}l@{}}$C_1$: Regularization parameter for the positive class\\ $C_2$: Regularization parameter for the negative class \\ $\sigma$: Kernel parameter \\ $\mu$: Intuitionistic fuzzy Kernel parameter \end{tabular}} &
  \multicolumn{1}{l|}{\begin{tabular}[c]{@{}l@{}}$C_1=[2^{-5},2^{-3},2^{-1},2^{1},2^3,2^5]$  \\ $C_2=[2^{-5},2^{-3},2^{-1},2^{1},2^3,2^5]$ \\ $\sigma=[2^{-10},2^{-9},\cdots,2^{10}]$ \\ $\mu=[2^{-10},2^{-9},\cdots,2^{10}]$\end{tabular}} \\ \hline
  \multicolumn{1}{|l|}{LSSVM-L} &
  \multicolumn{1}{l|}{\begin{tabular}[c]{@{}l@{}}$C$: Regularization parameter \end{tabular}} &
  \multicolumn{1}{l|}{\begin{tabular}[c]{@{}l@{}}$C=[2^{-5},2^{-3},2^{-1},2^{1},2^3,2^5]$ \end{tabular}} \\ \hline
\multicolumn{1}{|l|}{LSSVM-K} &
  \multicolumn{1}{l|}{\begin{tabular}[c]{@{}l@{}}$C$: Regularization parameter \\ $\sigma$: Kernel parameter \end{tabular}} &
  \multicolumn{1}{l|}{\begin{tabular}[c]{@{}l@{}}$C=[2^{-5},2^{-3},2^{-1},2^{1},2^3,2^5]$ \\ $\sigma=[2^{-10},2^{-9},\cdots,2^{10}]$ \end{tabular}} \\ \hline
   \multicolumn{1}{|l|}{LSTSVM-L} &
  \multicolumn{1}{l|}{\begin{tabular}[c]{@{}l@{}}$C_1$: Regularization parameter for the positive class\\ $C_2$: Regularization parameter for the negative class \end{tabular}} &
  \multicolumn{1}{l|}{\begin{tabular}[c]{@{}l@{}}$C_1=[2^{-5},2^{-3},2^{-1},2^{1},2^3,2^5]$  \\ $C_2=[2^{-5},2^{-3},2^{-1},2^{1},2^3,2^5]$ \end{tabular}} \\ \hline
  \multicolumn{1}{|l|}{LSTSVM-K} &
  \multicolumn{1}{l|}{\begin{tabular}[c]{@{}l@{}}$C_1$: Regularization parameter for the positive class\\ $C_2$: Regularization parameter for the negative class \\ $\sigma$: Kernel parameter \end{tabular}} &
  \multicolumn{1}{l|}{\begin{tabular}[c]{@{}l@{}}$C_1=[2^{-5},2^{-3},2^{-1},2^{1},2^3,2^5]$  \\ $C_2=[2^{-5},2^{-3},2^{-1},2^{1},2^3,2^5]$ \\ $\sigma=[2^{-10},2^{-9},\cdots,2^{10}]$ \end{tabular}} \\ \hline
  \multicolumn{1}{|l|}{Linex-SVM-L} &
  \multicolumn{1}{l|}{\begin{tabular}[c]{@{}l@{}}$C$: Regularization parameter \\ $a$: Linex loss function parameter\\ $t$: Initial value\\ $k$: Learning rate decay parameter\\
        $epsilon$:  Error tolerance \\
        $r$:  Momentum parameter\\
        $max\_it$: maximum iteration number\\
         $m$: Mini-batch size\\ \end{tabular}} &
  \multicolumn{1}{l|}{\begin{tabular}[c]{@{}l@{}}$C=[2^{-5},2^{-3},2^{-1},2^{1},2^3,2^5]$ \\ $a=-10:1:-1$\\ $t=0$\\ $k = 0.1$\\
        $epsilon = 10^{-8}$\\
        $r=0.6$\\
        $max\_it = 5000$\\
        $m=100$\\ \end{tabular}} \\ \hline
        \multicolumn{1}{|l|}{Linex-SVM-K} &
  \multicolumn{1}{l|}{\begin{tabular}[c]{@{}l@{}}$C$: Regularization parameter \\ $a$: Linex loss function parameter\\ $\sigma$: Kernel parameter \\ $t$: Initial value\\ $k$: Learning rate decay parameter\\
        $epsilon$:  Error tolerance \\
        $r$:  Momentum parameter\\
        $max\_it$: maximum iteration number\\
         $m$: Mini-batch size\\ \end{tabular}} &
  \multicolumn{1}{l|}{\begin{tabular}[c]{@{}l@{}}$C$: Regularization parameter \\ $a=-10:1:-1$\\ $\sigma=[2^{-10},2^{-9},\cdots,2^{10}]$ \\ $t=0$\\ $k = 0.1$\\
        $epsilon = 10^{-8}$\\
        $r=0.6$\\
        $max\_it = 5000$\\
         $m=100$\\ \end{tabular}} \\ \hline
\multicolumn{1}{|l|}{Pin-SVM-L} &
  \multicolumn{1}{l|}{\begin{tabular}[c]{@{}l@{}}$C$: Regularization parameter \\ $\tau$: Pinball loss function parameter \end{tabular}} &
  \multicolumn{1}{l|}{\begin{tabular}[c]{@{}l@{}}$C=[2^{-5},2^{-3},2^{-1},2^{1},2^3,2^5]$ \\ $\tau=[0,0.1,0.2,0.3,0.4,0.5,0.6,0.7,0.8,0.9,1]$ \end{tabular}} \\ \hline
\multicolumn{1}{|l|}{Pin-SVM-K} &
  \multicolumn{1}{l|}{\begin{tabular}[c]{@{}l@{}}$C$: Regularization parameter \\ $\sigma$: Kernel parameter\\ $\tau$: Pinball loss function parameter \end{tabular}} &
  \multicolumn{1}{l|}{\begin{tabular}[c]{@{}l@{}}$C=[2^{-5},2^{-3},2^{-1},2^{1},2^3,2^5]$ \\ $\sigma=[2^{-10},2^{-9},\cdots,2^{10}]$\\ $\tau=[0,0.1,0.2,0.3,0.4,0.5,0.6,0.7,0.8,0.9,1]$ \end{tabular}} \\ \hline
  \multicolumn{1}{|l|}{Pin-GTSVM-L} &
  \multicolumn{1}{l|}{\begin{tabular}[c]{@{}l@{}}$C_1$: Regularization parameter for the positive class\\ $C_2$: Regularization parameter for the negative class \\ $\tau_1$: Pinball loss function parameter for the positive class \\ $\tau_2$: Pinball loss function parameter for the negative class \end{tabular}} &
  \multicolumn{1}{l|}{\begin{tabular}[c]{@{}l@{}}$C_1=[2^{-5},2^{-3},2^{-1},2^{1},2^3,2^5]$ \\ $C_2=[2^{-5},2^{-3},2^{-1},2^{1},2^3,2^5]$ \\ $\tau_1=[0,0.1,0.2,0.3,0.4,0.5,0.6,0.7,0.8,0.9,1]$ \\ $\tau_2=[0,0.1,0.2,0.3,0.4,0.5,0.6,0.7,0.8,0.9,1]$ \end{tabular}} \\ \hline
\multicolumn{1}{|l|}{Pin-GTSVM-K} &
  \multicolumn{1}{l|}{\begin{tabular}[c]{@{}l@{}}$C_1$: Regularization parameter for the positive class\\ $C_2$: Regularization parameter for the negative class \\ $\sigma$: Kernel parameter\\ $\tau_1$: Pinball loss function parameter for the positive class \\ $\tau_2$: Pinball loss function parameter for the negative class \end{tabular}} &
  \multicolumn{1}{l|}{\begin{tabular}[c]{@{}l@{}}$C_1=[2^{-5},2^{-3},2^{-1},2^{1},2^3,2^5]$ \\ $C_2=[2^{-5},2^{-3},2^{-1},2^{1},2^3,2^5]$ \\ $\sigma=[2^{-10},2^{-9},\cdots,2^{10}]$\\ $\tau_1=[0,0.1,0.2,0.3,0.4,0.5,0.6,0.7,0.8,0.9,1]$ \\ $\tau_2=[0,0.1,0.2,0.3,0.4,0.5,0.6,0.7,0.8,0.9,1]$ \end{tabular}} \\ \hline
\end{tabular}
}
\end{table}
%%%%%%%%%%%%%%%%%%%%%%%%%
\begin{table}[ht!]
\centering
\caption{Index of activation functions used in the RNN-based classifiers. }
\label{tab:activation_function}
\begin{tabular}{|c|l|}
\hline
\textbf{Index} & \multicolumn{1}{c|}{\textbf{Activation Functions}}    \\ \hline
1              & Scaled Exponential Linear Unit (SELU)                 \\ \hline
2              & Rectified Linear Units (ReLU)                         \\ \hline
3              & Sigmoid                                               \\ \hline
4              & Sine (Sin)                                            \\ \hline
5              & Hard Limit Transfer Function (Hardlim)                \\ \hline
6              & Triangular Basis Transfer Function (Tribas)           \\ \hline
7              & Radial Basis Transfer Function (Radbas)               \\ \hline
8              & Signum (Sgn)                                          \\ \hline
9              & Hyperbolic Tangent Sigmoid Transfer Function (Tansig) \\ \hline
\end{tabular}
\end{table}
%%%%%%%%%%%%%%%%%%%%%
\subsubsection{Implementation Details and Experimental Setup}
The experimental procedures are executed on a computing system possessing MATLAB R2023a software, Intel(R) Xeon(R) Platinum 8260 CPU @ 2.30GHz, 2301 Mhz, 24 Core(s), 48 Logical Processor(s) with 256 GB RAM on a Windows-10 operating platform. The dataset is split into a $70\%$ portion for model training and a $30\%$ portion for testing. We employ the 5-fold cross-validation approach and grid search in the training data to fine-tune model hyperparameters. In this, one subset of the dataset is set aside for validation in each iteration, while the remaining four are used for training. We estimated the validation accuracy for each fold independently for each set of hyperparameters. The mean of these five accuracies was then used to get the average validation accuracy for each set of hyperparameters. The best hyperparameters are selected with the highest average validation accuracy. Then, the Accuracy (Acc.), Sensitivity (Sens.), Specificity (Spec.), Precision (Prec.), and F-measure (f-measure) are determined using the best hyperparameters on the testing datasets. 
\subsubsection{Hyperparameters' Settings} 
To get the best hyperparameter settings for our experiments, we applied a 5-fold cross-validation-based grid search technique. Each experiment is run $20$ times to get the robust and best solution. The detailed hyperparameters' description and their ranges for RNN and HbCs employed in the grid search are comprehensively presented in Tables \ref{tab:Randomized_Parameters_Range} and \ref{tab:hyperplane_Parameters_Range}, respectively. A total of $9$ activation functions are systematically explored and fine-tuned during the grid search experiment of RNN-based classifiers. The activation functions and their corresponding indices utilized in this study are documented in Table \ref{tab:activation_function}. Gaussian Kernel is employed for the Kernelized HbCs.
%%%%%%%%%%%%%%%%%%%%%%%%%%%
\subsubsection{Performance Metrics} This work primarily focused on accurately diagnosing SMC using extracted features from MRI scans. To assess its effectiveness and to provide a comprehensive comparison, the algorithm's performance metrics Accuracy (Acc.), Sensitivity(Sens.), Specificity (Spec.), Precision (Prec.), and F-measure (f-measure) are meticulously compared against established classifier models. 

The formulae of statistical metrics are used to assess the algorithms are as follows:
\begin{align}
  \text{Acc. ~(Accuracy)}&= \frac{T_+ + T_-}{T_+ + F_+ + T_- +F_-},\\
  \text{Sens. ~(Sensitivity)}&= \frac{T_+}{T_+ +F_-},\\
  \text{Spec. ~(Specificity)}&= \frac{T_-}{T_- +F_+},\\
  \text{Prec. ~(Precision)}&= \frac{T_+}{T_+ +F_+},\\
  \text{F-measure (f-measure)}&= 2 \times \frac{ Prec. \times Sens.}{Prec. + Sens.},
\end{align}
the terms false positive ($F_+$), true positive ($T_+$), false negative ($F_-$), and true negative ($T_-$) are used to represent different outcomes.

The Acc. metric comprehensively evaluates the system's overall proficiency and the classifier's adeptness in discerning between SMC and Healthy Control (HC) cases. Sens. measures the ratio of true positives to all positives in the confusion matrix (CM), while Spec. gauges the model's capability in predicting true negatives. Prec quantifies the accuracy of positive predictions relative to all positive predictions. The f-measure quantifies the model's efficacy on the dataset. Using CM as a visual aid for evaluating classification effectiveness transcends conventional accuracy assessments by illustrating correct and incorrect predictions for each class.
%%%%%%%%%%%%%%%%%%%%%%%%%%%%%%%
\subsection{Comparison of Different State-of-the-art RNN-based Classifiers and Statistical Analysis}  
In this comprehensive analysis, we scrutinize the performance of $13$ RNN-based classifiers for the binary classification task distinguishing between HC subjects and those with SMC. Through rigorous examination, this research aims to illuminate the optimal RNN-based model for accurate and reliable binary classification in the context of cognitive health assessments over the $5$ feature sets, \textit{i.e.}, CT, GM, WM, JD, and all feature sets combined, encompassing a diverse range of demographic and clinical profiles. Experimental results corresponding to all the performance metrics are reported in Table \ref{rnntab}. The best hyperparameters corresponding to each model for every modality are reported in Table \ref{tab:Best_Parameters_RNN}. Among the examined RNN-based classifiers, the dRVFL exhibits the preeminence with the highest average accuracy at $75.76\%$, closely followed by the edRVFL, achieving an accuracy of $73.94\%$. Across multiple modalities, namely CT, GM, WM, JD, and the fusion of all feature sets, the dRVFL and edRVFL consistently surpass their RNN counterparts in terms of testing accuracy. Notably, the competing RNN classifiers struggle to breach the $65\%$ accuracy threshold. The dRVFL and edRVFL models demonstrate their effectiveness by consistently achieving an average accuracy approximately $10\%$ to $16\%$ higher than other models. In terms of sensitivity, the dRVFL model leads the pack with a commanding value of $79.22\%$. The edRVFL model distinguishes itself in terms of specificity, precision, and f-measure, boasting the highest values of $75.67\%$, $78.91\%$, and $75.67\%$, respectively. The discernible advantages exhibited by dRVFL and edRVFL underscore their superior performance, positioning them as the best choices for the HC vs. SMC task. 

%%%%%%%%%%%%%%%%%%%%%%%% 
\begin{table*}[ht!]
\centering
\caption{The performance of RNN-based classifiers \textit{w.r.t.} all the modalities, \textit{i.e.}, CT, GM, JD, WM, and All Features.} \label{rnntab}
\resizebox{\linewidth}{!}{%
\begin{tabular}{lllllll}
\hline
\textbf{RNN} & \multicolumn{1}{c}{\textbf{CT}} & \multicolumn{1}{c}{\textbf{GM}} & \multicolumn{1}{c}{\textbf{JD}} & \multicolumn{1}{c}{\textbf{WM}} & \multicolumn{1}{c}{\textbf{All Features}} & \multicolumn{1}{c}{\textbf{Average}} \\ 
\hline
\textbf{RVFL} & {[}65.16, 61.3, 68.58, 63.34, 62.3{]} & {[}65.16, 64.71, 65.63, 66.67, 65.68{]} & {[}57.58, 61.3,   54.29, 54.29, 57.58{]} & {[}66.67, 73.08,   62.5, 55.89, 63.34{]} & {[}71.22, 80, 61.3, 70, 74.67{]} & {[}65.15, 68.07, 62.45, 62.03, 64.71{]}\\[0.3cm]

\textbf{ELM} & {[}57.58, 62.5, 52.95, 55.56, 58.83{]} & {[}59.1, 62.86, 54.84, 61.12, 61.98{]} & {[}60.61, 54.84, 65.72, 58.63, 56.67{]} & {[}59.1, 65.72, 51.62, 60.53, 63.02{]} & {[}63.64, 77.42, 51.43, 58.54, 66.67{]} & {[}60, 64.67, 55.31, 58.87, 61.43{]} \\[0.3cm]

\textbf{MCVELM} & {[}62.13, 56.67, 66.67, 58.63, 57.63{]} & {[}60.61, 62.86, 58.07, 62.86, 62.86{]} & {[}60.61, 58.34, 63.34, 65.63, 61.77{]} & {[}63.64, 65.52, 62.17, 57.58, 61.3{]} & {[}63.64, 58.83, 68.75, 66.67, 62.5{]} & {[}62.12, 60.44, 63.8, 62.27, 61.21{]}\\[0.3cm]

\textbf{MVELM} & {[}62.13, 65.72, 58.07, 63.89, 64.79{]} & {[}68.19, 64.87, 72.42, 75, 69.57{]} & {[}59.1, 48.84, 78.27, 80.77, 60.87{]} & {[}60.61, 54.55, 66.67, 62.07, 58.07{]} & {[}65.16, 64.71, 65.63, 66.67, 65.68{]} & {[}63.03, 59.73, 68.21, 69.68, 63.79{]}\\[0.3cm]

\textbf{IFRVFL} & {[}62.13, 60.61, 63.64, 62.5, 61.54{]} & {[}60.61, 56.25, 64.71, 60, 58.07{]} & {[}65.16, 72.5, 53.85, 70.74, 71.61{]} & {[}62.13, 54.55, 69.7, 64.29, 59.02{]} & {[}56.07, 45.72, 67.75, 61.54, 52.46{]} & {[}61.21, 57.92, 63.93, 63.81, 60.54{]}\\[0.3cm]

\textbf{Class-Var-RVFL} & {[}57.58, 62.5, 52.95, 55.56, 58.83{]} & {[}59.1, 64.11, 51.86, 65.79, 64.94{]} & {[}60.61, 62.07, 59.46, 54.55, 58.07{]} & {[}62.13, 64.52, 60, 58.83, 61.54{]} & {[}56.07, 60.61, 51.52, 55.56, 57.98{]} & {[}59.09, 62.76, 55.15, 58.05, 60.27{]}\\[0.3cm]

\textbf{Total-Var-RVFL} & {[}60.61, 53.58, 65.79, 53.58, 53.58{]} & {[}57.58, 62.86, 51.62, 59.46, 61.12{]} & {[}62.13, 64.52, 60, 58.83, 61.54{]} & {[}56.07, 54.06, 58.63, 62.5, 57.98{]} & {[}60.61, 100, 0, 60.61, 75.48{]} & {[}59.39, 67, 47.2, 58.99, 61.93{]}\\[0.3cm]

\textbf{GEELM-LDA} & {[}60.61, 55.27, 67.86, 70, 61.77{]} & {[}60.61, 82.15, 44.74, 52.28, 63.89{]} & {[}65.16, 71.06, 57.15, 69.24, 70.13{]} & {[}62.13, 60.61, 63.64, 62.5, 61.54{]} & {[}59.1, 12, 87.81, 37.5, 18.19{]} & {[}61.52, 56.21, 64.24, 58.3, 55.1{]} \\[0.3cm]

\textbf{GEELM-LFDA} & {[}57.58, 53.13, 61.77, 56.67, 54.84{]} & {[}65.16, 62.86, 67.75, 68.75, 65.68{]} & {[}66.67, 54.55, 78.79, 72, 62.07{]} & {[}60.61, 64.52, 57.15, 57.15, 60.61{]} & {[}59.1, 65.52, 54.06, 52.78, 58.47{]} & {[}61.82, 60.11, 63.9, 61.47, 60.33{]} \\[0.3cm]

\textbf{dRVFL} & {[}77.28, 77.78, 76.67, 80, 78.88{]} & {[}71.22, 76.48, 65.63, 70.28, 73.24{]} & {[}74.25, 81.49, 69.24, 64.71, 72.14{]} & {[}77.28, 75.76, 78.79, 78.13, 76.93{]} & {[}78.79, 84.62, 75, 68.75, 75.87{]} & {[}\textbf{\underline{75.76}}, \textbf{\underline{79.22}}, 73.06, 72.37, 75.41{]} \\[0.3cm]

\textbf{edRVFL} & {[}78.79, 78.95, 78.58, 83.34, 81.09{]} & {[}74.25, 69.45, 80, 80.65, 74.63{]} & {[}69.7, 65.91, 77.28, 85.3, 74.36{]} & {[}74.25, 74.36, 74.08, 80.56, 77.34{]} & {[}72.73, 78.58, 68.43, 64.71, 70.97{]} & {[}73.94, 73.45, \textbf{\underline{75.67}}, \textbf{\underline{78.91}}, \textbf{\underline{75.67}}{]} \\[0.3cm]

\textbf{BLS} & {[}62.13, 64.52, 60, 58.83, 61.54{]} & {[}59.1, 65.63, 52.95, 56.76, 60.87{]} & {[}54.55, 54.55, 54.55, 54.55, 54.55{]} & {[}60.61, 44.74, 82.15, 77.28, 56.67{]} & {[}60.61, 64.87, 55.18, 64.87, 64.87{]} & {[}59.39, 58.86, 60.96, 62.45, 59.7{]} \\[0.3cm]

\textbf{NF-BLS} & {[}54.55, 66.67, 47.62, 42.11, 51.62{]} & {[}62.13, 61.12, 63.34, 66.67, 63.77{]} & {[}57.58, 45.72, 70.97, 64, 53.34{]} & {[}54.55, 64.71, 43.75, 55, 59.46{]} & {[}65.16, 64.52, 65.72, 62.5, 63.5{]} & {[}58.79, 60.54, 58.28, 58.05, 58.33{]}\\[0.3cm]
\hline 
\textbf{Average} & {[}62.94, 63.01, 63.16, 61.84, 62.09{]} & {[}63.29, 65.86, 61.04, 65.1, \textbf{\underline{65.1}}{]} & {[}62.59, 61.2, \textbf{\underline{64.83}}, \textbf{\underline{65.63}}, 62.67{]} & {[}63.05, 62.82, 63.91, 64.02, 62.83{]} & {[}\textbf{\underline{63.99}}, \textbf{\underline{65.95}}, 59.42, 60.82, 62.1{]} \\[0.1cm]
\hline \\[-0.4cm]
\end{tabular}%
}
\footnotesize{The above performance is in \% and is mentioned in the format as follows: [Acc., Sens., Spec., Prec., f-measure].}
\footnotesize{The last column of the table is the average performance of each model for all the modalities. Boldface and underline in the last column refer to the best-performed model \textit{w.r.t.} each modality.}
\footnotesize{The last row of the table is the average performance contributed by each modality to the classification for all the models. Boldface and underline in the last row refer to the most prominent modality \textit{w.r.t.} Acc., Sens., Spec., Prec., and f-measure.}\\
\end{table*}
% \usepackage{graphicx}
\begin{table}[]
\centering
\caption{Best parameters based on grid search technique for RNN-based classifiers.}
\label{tab:Best_Parameters_RNN}
\resizebox{\columnwidth}{!}{%
\begin{tabular}{|l|l|l|l|l|l|l|}
\hline
\multicolumn{1}{|c|}{\textbf{}} &
  \multicolumn{1}{c|}{\textbf{Parameters Sequence}} &
  \multicolumn{1}{c|}{\textbf{CT}} &
  \multicolumn{1}{c|}{\textbf{GM}} &
  \multicolumn{1}{c|}{\textbf{JD}} &
  \multicolumn{1}{c|}{\textbf{WM}} &
  \multicolumn{1}{c|}{\textbf{All Feature}} \\ \hline
\textbf{RVFL} &
  ($Act$, $C$, $N$) &
  (9, 1000000, 323) &
  (1, 1000000, 203) &
  (6, 100000000, 463) &
  (1, 1000000, 183) &
  (2, 100000000, 323) \\ \hline
\textbf{ELM} &
  ($Act$, $C$, $N$) &
  (9, 100, 103) &
  (3, 100000000, 83) &
  (2, 100, 23) &
  (3, 100, 383) &
  (7, 1000000, 183) \\ \hline
\textbf{MCVELM} &
  ($Act$, $C$, $\lambda$, $N$) &
  (2, 10000, 0.000001, 363) &
  (2, 100000000, 1000000, 183) &
  (1, 1000000, 10000, 323) &
  (8, 1000000, 10000, 443) &
  (3, 100, 0, 363) \\ \hline
\textbf{MCVELM} &
  ($Act$, $C$, $\lambda$, $N$) &
  (7, 100, 1, 103) &
  (4, 0.01, 0, 463) &
  (1, 1, 0.01, 323) &
  (2, 1000000, 1, 243) &
  (9, 1000000, 0.0001, 283) \\ \hline
\textbf{IFRVFL} &
  ($Act$, $C$, $\mu$, $N$) &
  (1, 100000000, 0.125, 63) &
  (7, 10000, 4, 463) &
  (8, 100, 0.25, 223) &
  (1, 10000, 0.125, 83) &
  (8, 10000, 0.5, 223) \\ \hline
\textbf{Class-Var-RVFL} &
  ($Act$, $C$, $\lambda$, $N$) &
  (5, 1000000, 0, 383) &
  (4, 1000000, 1000000, 303) &
  (3, 1, 0.000001, 163) &
  (1, 100, 100, 143) &
  (8, 1000000, 100, 183) \\ \hline
\textbf{Total-Var-RVFL} &
  ($Act$, $C$, $\lambda$, $N$) &
  (9, 100, 100, 483) &
  (5, 1000000, 0.000001, 203) &
  (2, 100000000, 10000, 23) &
  (7, 1000000, 0.000001, 383) &
  (6, 10000, 100, 303) \\ \hline
\textbf{GEELM-LDA} &
  ($Act$, $C$, $\lambda$, $N$) &
  (6, 100000000, 0, 403) &
  (3, 100000000, 0.000001, 403) &
  (4, 100, 0, 63) &
  (7, 10000, 0.0001, 43) &
  (3, 1000000, 0.01, 223) \\ \hline
\textbf{GEELM-LFDA} &
  ($Act$, $C$, $\lambda$, $N$) &
  (5, 100000000, 1000000, 383) &
  (4, 1000000, 1, 123) &
  (3, 100000000, 1, 203) &
  (6, 1, 0, 503) &
  (4, 100000000, 100, 323) \\ \hline
\textbf{dRVFL} &
  ($Act$, $C$, $N$, $L$) &
  (1, 0.015469, 226, 3) &
  (7, 0.012656, 973, 8) &
  (6, 0.147623, 200, 9) &
  (8, 0.010252, 730, 5) &
  (9, 3.24, 276, 4) \\ \hline
\textbf{edRVFL} &
  ($Act$, $C$, $N$, $L$) &
  (3, 0.049359, 534, 3) &
  (6, 0.015625, 1024, 3) &
  (8, 0.155824, 243, 3) &
  (1, 3.24, 564, 3) &
  (5, 10.4976, 283, 7) \\ \hline
\textbf{BLS} &
  ($Act$, $C$, $N_{Feat-G}$, $N_{Feat-N}$, $N_{E-G}$) &
  (4, 1000000, 5, 15, 15) &
  (2, 10000, 30, 7, 45) &
  (1, 1000000, 15, 15, 85) &
  (1, 100, 40, 7, 75) &
  (8, 0.000001, 20, 1, 15) \\ \hline
\textbf{NF-BLS} &
  ($Act$, $C$, $N_{Fuzz-G}$, $N_{Fuzz-N}$, $N_{E-G}$) &
  (7, 0.00000001, 3, 8, 73) &
  (4, 0.00000001, 33, 5, 83) &
  (4, 0.000001, 13, 8, 13) &
  (1, 1, 13, 1, 23) &
  (5, 0.0001, 73, 7, 43) \\ \hline
\end{tabular}%
}
\end{table}
% \usepackage{graphicx}
% \usepackage[normalem]{ulem}
% \usepackage{lscape}
% \begin{landscape}
\begin{table}[]
\centering
\caption{Rankings for RNN-based models with respect to accuracy.}
\label{tab:Rank_RNN}
\resizebox{\columnwidth}{!}{%
\begin{tabular}{|l|c|c|c|c|c|c|c|c|c|c|c|c|c|}
\hline
\textbf{} &
  \multicolumn{1}{l|}{\textbf{RVFL}} &
  \multicolumn{1}{l|}{\textbf{ELM}} &
  \multicolumn{1}{l|}{\textbf{MCVELM}} &
  \multicolumn{1}{l|}{\textbf{MVELM}} &
  \multicolumn{1}{l|}{\textbf{IFRVFL}} &
  \multicolumn{1}{l|}{\textbf{Class-Var-RVFL}} &
  \multicolumn{1}{l|}{\textbf{Total-var-RVFL}} &
  \multicolumn{1}{l|}{\textbf{GEELM-LDA}} &
  \multicolumn{1}{l|}{\textbf{GEELM-LFDA}} &
  \multicolumn{1}{l|}{\textbf{dRVFL}} &
  \multicolumn{1}{l|}{\textbf{edRVFL}} &
  \multicolumn{1}{l|}{\textbf{BLS}} &
  \multicolumn{1}{l|}{\textbf{NF-BLS}} \\ \hline
\textbf{CT}          & 3    & 11  & 5.5 & 5.5 & 5.5  & 11   & 8.5 & 8.5  & 11   & 2                  & 1   & 5.5 & 13   \\ \hline
\textbf{GM}          & 4.5  & 11  & 8   & 3   & 8    & 11   & 13  & 8    & 4.5  & 2                  & 1   & 11  & 6    \\ \hline
\textbf{JD}          & 11.5 & 8   & 8   & 10  & 4.5  & 8    & 6   & 4.5  & 3    & 1                  & 2   & 13  & 11.5 \\ \hline
\textbf{WM}          & 3    & 11  & 4   & 9   & 6    & 6    & 12  & 6    & 9    & 1                  & 2   & 9   & 13   \\ \hline
\textbf{All Feature} & 3    & 6.5 & 6.5 & 4.5 & 12.5 & 12.5 & 8.5 & 10.5 & 10.5 & 1                  & 2   & 8.5 & 4.5  \\ \hline
\textbf{Average}     & 5    & 9.5 & 6.4 & 6.4 & 7.3  & 9.7  & 9.6 & 7.5  & 7.6  & {\ul \textbf{1.4}} & 1.6 & 9.4 & 9.6  \\ \hline
\end{tabular}%
}
\end{table}
% \end{landscape}
%%%%%%%%%%%%%%%%%%%%

To discern the predominant impact of different modalities on the HC vs. SMC classification task, we computed the average values across all models. The analysis, gleaned from the last row of Table \ref{rnntab}, reveals that, from the standpoint of accuracy and sensitivity, the All feature set commands a pivotal role, achieving percentages of $63.99\%$ and $65.95\%$, respectively. Notably, JD emerges as superior in feature representation among modalities, boasting $64.83\%$ and $65.63\%$ for specificity and precision, respectively. In terms of the f-measure, the GM modality assumes the utmost importance, registering the highest value at $65.1\%$. CM for the dRVFL model for the GM modality has been visualized in Figure \ref{conf}(a). 

Average accuracy, while a common metric, can be misleading by masking a model's varied performance across diverse datasets. In response, we adhere to Demsar \cite{demvsar2006statistical} counsel and implement statistical tests, such as the ranking test, Friedman test, and Win-tie-loss sign test. This rigorous approach facilitates a thorough evaluation of model performance, mitigating bias and empowering us to draw extensive conclusions regarding the models' efficacy. In the ranking scheme, models are evaluated based on their performance across individual datasets. Higher ranks are assigned to models with poorer performance, while top performers receive lower ranks. Suppose we are assessing $\mathscr{N}$ models across $\mathscr{P}$ datasets. The rank of the $n^{th}$ model on the $p^{th}$ dataset is denoted as $\mathscr{R}_{n}^p$. The average rank of the $n^{th}$ model is computed as follows: $\mathscr{R}_{n}=\left(\sum_{p=1}^{\mathscr{P}}\mathscr{R}_{n}^p\right)/\mathscr{P}$. The rank of each model $\textit{w.r.t.}$ each modality is reported in Table \ref{tab:Rank_RNN}. The table's last row indicates the average rank for each model. The dRVFL and edRVFL secure the top two positions at $1.4$ and $1.6$, respectively. In contrast, Class-Var-RVFL lags with an average rank of $9.7$. Lower ranks signify higher model superiority, highlighting the distinct superiority of dRVFL and edRVFL. We utilize the Friedman test \cite{demvsar2006statistical} to compare the average ranks of models and determine if there are significant differences in their rankings. This statistical test relies on a chi-squared distribution ($\chi^2_F$) with $\mathscr{N}-1$ degrees of freedom (dof). The formula for Friedman's test is given as: 
$\chi^2_F = \frac{12\mathscr{P}}{\mathscr{N} (\mathscr{N}+1)} \left(\sum_{n=1}^{\mathscr{N}} \left(\mathscr{R}_{n}\right)^2 - \frac{\mathscr{N}(\mathscr{N}+1)^2}{4}\right).$ The $F_F$ statistic is computed as follows: $ F_F=\chi_F^2\left(\frac{(\mathscr{P}-1)}{\mathscr{P}(\mathscr{N}-1)-\chi_F^2}\right).$ The distribution of $F_F$ is characterized by $(\mathscr{N}-1)$ and $(\mathscr{P}-1)(\mathscr{N}-1)$ dof. For $\mathscr{N}=13$ and $\mathscr{P}=5$, we obtain $\chi^2_F=32.56$ and $F_F=4.75$. Referring to the statistical table for the $F$-distribution, we find that $F_F (12, 48) = 1.96$ at a significance level of $5\%$. The null hypothesis is rejected because $4.75 > 1.96$. Friedman test indicates significant differences among the compared SOTA RNN-based classifiers for the HC vs. SMC classification task. 

Further, we discuss the Win-tie-loss sign test for the RNN-based classifiers. The win-tie-loss sign test \cite{demvsar2006statistical} is a widely utilized statistical method in research and data analysis to assess whether there exists a statistically significant difference between the outcomes of two or more models. According to the null hypothesis of this test, two models are considered to exhibit equivalent performance if each wins on $\mathscr{P}/2$ of the $\mathscr{P}$ datasets. For two models to be deemed significantly different, each must secure at least $\mathscr{P}/2 + 1.96\sqrt{\mathscr{P}}/2$ wins. In cases of an even number of ties, these are evenly distributed between the models. However, when ties are odd, one tie is excluded, and the remaining ties are uniformly allocated between the models. In our case $\mathscr{P}$= $5$, the threshold for determining statistical difference according to the win-tie-loss test is equal to $\mathscr{P}/2 + 1.96\sqrt{\mathscr{P}}/2$ = $5/2 + 1.96\sqrt{5}/2$ = $4.69$. 

% \usepackage{graphicx}
% \usepackage{lscape}
% \begin{landscape}
\begin{table}[ht!]
\centering
\caption{Pairwise Win-tie-loss for RNN-based classifiers.}
\label{tab:Win-Tie-Loss_RNN}
% \resizebox{\columnwidth}{!}{%
 \resizebox{18cm}{!}{
\begin{tabular}{|l|c|c|c|c|c|c|c|c|c|c|c|c|}
\hline
 &
  RVFL &
  ELM &
  MCVELM &
  MVELM &
  IFRVFL &
  Class-Var-RVFL &
  Total-var-RVFL &
  GEELM-LDA &
  GEELM-LFDA &
  dRVFL &
  edRVFL &
  BLS \\ \hline
ELM &
  [$1, 0, 4$] &
   &
   &
   &
   &
   &
   &
   &
   &
   &
   &
   \\ \hline
MCVELM &
  [$1, 0, 4$] &
  [$3, 2, 0$] &
   &
   &
   &
   &
   &
   &
   &
   &
   &
   \\ \hline
MVELM &
  [$2, 0, 3$] &
  [$4, 0, 1$] &
  [$2, 1, 2$] &
   &
   &
   &
   &
   &
   &
   &
   &
   \\ \hline
IFRVFL &
  [$1, 0, 4$] &
  [$4, 0, 1$] &
  [$1, 2, 2$] &
  [$2, 1, 2$] &
   &
   &
   &
   &
   &
   &
   &
   \\ \hline
Class-Var-RVFL &
  [$1, 0, 4$] &
  [$1, 3, 1$] &
  [$0, 1, 4$] &
  [$2, 0, 3$] &
  [$0, 2, 3$] &
   &
   &
   &
   &
   &
   &
   \\ \hline
Total-var-RVFL &
  [$1, 0, 4$] &
  [$2, 0, 3$] &
  [$1, 0, 4$] &
  [$1, 0, 4$] &
  [$1, 0, 4$] &
  [$3, 0, 2$] &
   &
   &
   &
   &
   &
   \\ \hline
GEELM-LDA &
  [$1, 0, 4$] &
  [$4, 0, 1$] &
  [$1, 1, 3$] &
  [$2, 0, 3$] &
  [$1, 3, 1$] &
  [$4, 1, 0$] &
  [$3, 1, 1$] &
   &
   &
   &
   &
   \\ \hline
GEELM-LFDA &
  [$1, 1, 3$] &
  [$3, 1, 1$] &
  [$2, 0, 3$] &
  [$1, 1, 3$] &
  [$3, 0, 2$] &
  [$3, 1, 1$] &
  [$3, 0, 2$] &
  [$2, 1, 2$] &
   &
   &
   &
   \\ \hline
dRVFL &
  [$5, 0, 0$] &
  [$5, 0, 0$] &
  [$5, 0, 0$] &
  [$5, 0, 0$] &
  [$5, 0, 0$] &
  [$5, 0, 0$] &
  [$5, 0, 0$] &
  [$5, 0, 0$] &
  [$5, 0, 0$] &
   &
   &
   \\ \hline
edRVFL &
  [$5, 0, 0$] &
  [$5, 0, 0$] &
  [$5, 0, 0$] &
  [$5, 0, 0$] &
  [$5, 0, 0$] &
  [$5, 0, 0$] &
  [$5, 0, 0$] &
  [$5, 0, 0$] &
  [$5, 0, 0$] &
  [$2, 0, 3$] &
   &
   \\ \hline
BLS &
  [$0, 0, 5$] &
  [$2, 1, 2$] &
  [$0, 1, 4$] &
  [$0, 2, 3$] &
  [$1, 1, 3$] &
  [$2, 1, 2$] &
  [$3, 1, 1$] &
  [$2, 0, 3$] &
  [$2, 1, 2$] &
  [$0, 0, 5$] &
  [$0, 0, 5$] &
   \\ \hline
NF-BLS &
  [$0, 1, 4$] &
  [$2, 0, 3$] &
  [$2, 0, 3$] &
  [$0, 1, 4$] &
  [$2, 0, 3$] &
  [$2, 0, 3$] &
  [$2, 0, 3$] &
  [$2, 0, 3$] &
  [$1, 0, 4$] &
  [$0, 0, 5$] &
  [$0, 0, 5$] &
  [$3, 0, 2$] \\ \hline 
\multicolumn{9}{l}{In the $[\alpha,\beta,\gamma]$, $\alpha$, $\beta$, and $\gamma$ denote the number of win, tie, and loss of the row model over the corresponding column model.}\\
\end{tabular}%
}
\end{table}
% \end{landscape}
%%%%%%%%%%%%%%%%%

The pairwise Win-tie-loss for the RNN-based classifiers' pairwise Win-tie-loss results are outlined in Table \ref{tab:Win-Tie-Loss_RNN}. Examining the table reveals that dRVFL and edRVFL exhibit 5-5 wins against other RNN-based classifiers, establishing their significant superiority over the rest of the models. No other classifier demonstrates a statistically significant difference. This statistical analysis underscores the superiority of dRVFL and edRVFL in the HC vs. SMC classification task compared to other state-of-the-art (SOTA) RNN-based classifiers. 
%%%%%%%%%%%%%%%%%%%%%
\subsection{Comparison of Different State-of-the-art HbCs and Statistical Analysis}      
\begin{figure*}[ht!]
    \centering
    \includegraphics[width=0.5\linewidth]{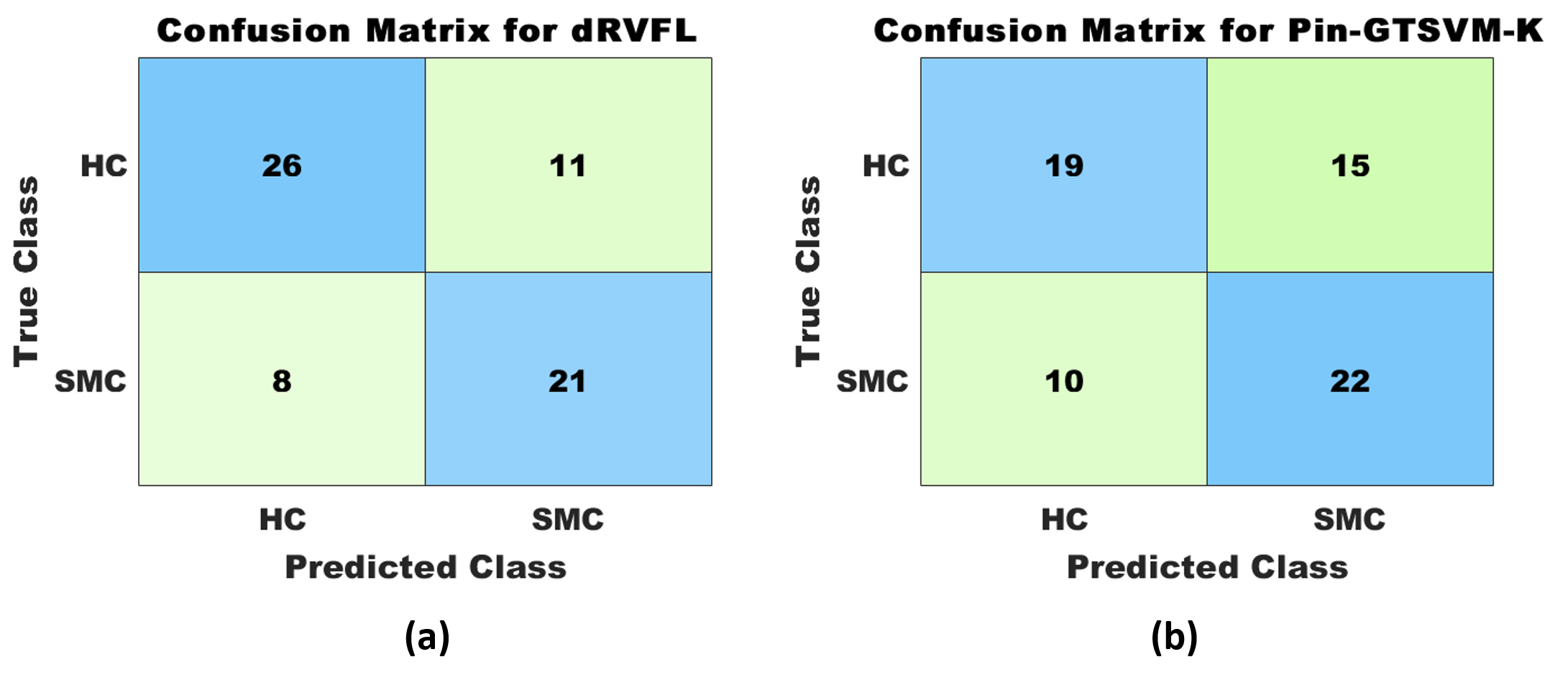}
    \caption{Confusion matrix for the top performing model of RNN and HbC, \textit{i.e.}, dRVL and Pin-GTSVM-K respectively.}
    \label{conf}
\end{figure*}

In this investigation, a meticulous evaluation is conducted to gauge the efficacy of $16$ HbCs in the binary classification task, discerning between HC subjects and those exhibiting SMC. The comprehensive set of performance metrics is delineated in Table \ref{svmtab}, while the best hyperparameters for each HbC across modalities are detailed in Table \ref{tab:Best_Parametrs_HbC}. Upon scrutiny of the last column in Table \ref{svmtab}, it is evident that the Pin-GTSVM-K model attains the highest average of $64.55\%$ among the HbCs. Regarding sensitivity, the Pin-SVM-K stands out with the highest value at $93.51\%$. In specificity, precision, and f-measure, the TSVM-K, SVM-L, and Linex-SVM-K exhibit supremacy with values of $72\%$, $70.56\%$, and $72.76\%$, respectively. Notably, a discernible trend emerges, indicating that most of the time, Kernel-based models outperform their linear counterparts in the HC vs. SMC classification. Considering the preeminent modality, scrutiny of the last row in the table reveals that the WM modality stands out with the highest values for average accuracy, sensitivity, precision, and f-measure at $64.3\%$, $73.31\%$, $63.82\%$, and $67.12\%$, respectively. In terms of specificity, the GM modality takes precedence, boasting the highest value at $54.35\%$. Consequently, in the realm of HbCs, the WM modality assumes a pivotal role in the classification of HC vs. SMC. CM for the Pin-GTSVM-K model for the GM modality has been visualized in Figure \ref{conf}(b). 
%%%%%%%%%%%%%%%%%%%%%%%%%%%%%%%%%%%%
\begin{table*}[]
\centering
\caption{The performance of HbCs \textit{w.r.t.} all the modalities, \textit{i.e.}, CT, GM, JD, WM, and All Features.} \label{svmtab}
\resizebox{\linewidth}{!}{%
\begin{tabular}{lllllll}
\hline
\textbf{HbC} & \multicolumn{1}{c}{\textbf{CT}} & \multicolumn{1}{c}{\textbf{GM}} & \multicolumn{1}{c}{\textbf{JD}} & \multicolumn{1}{c}{\textbf{WM}} & \multicolumn{1}{c}{\textbf{All Features}} & \multicolumn{1}{c}{\textbf{Average}} \\ 
\hline
\textbf{SVM-L} & {[}63.64, 67.57, 58.63, 67.57, 67.57{]} 
& {[}57.58, 55.27, 60.72, 65.63, 60{]}
 & {[}60.61, 57.9, 64.29, 68.75, 62.86{]}
 & {[}69.7, 62.86, 77.42, 75.87, 68.75{]}
 & {[}63.64, 55.27, 75, 75, 63.64{]} & {[}63.03, 59.77, 67.21, \textbf{\underline{70.56}}, 64.56{]} \\[0.3cm]

\textbf{SVM-K} & {[}63.64, 97.23, 23.34, 60.35, 74.47{]}
 & {[}57.58, 100, 0, 57.58, 73.08{]}
 & {[}51.52, 65.72, 35.49, 53.49, 58.98{]}
 & {[}60.61, 64.52, 57.15, 57.15, 60.61{]}
 & {[}56.07, 76.48, 34.38, 55.32, 64.2{]} & {[}57.88, 80.78, 30.07, 56.77, 66.26{]}
 \\[0.3cm]

\textbf{TSVM-L} & {[}62.13, 87.88, 36.37, 58, 69.88{]}
 & {[}59.1, 56.67, 61.12, 54.84, 55.74{]}
 & {[}60.61, 68, 56.1, 48.58, 56.67{]}
& {[}63.64, 67.75, 60, 60, 63.64{]}
 & {[}62.13, 69.7, 54.55, 60.53, 64.79{]} &{[}61.52, 70, 53.62, 56.39, 62.14{]}
 \\[0.3cm]

\textbf{TSVM-K} & {[}66.67, 72.73, 60.61, 64.87, 68.58{]}
 & {[}62.13, 40.63, 82.36, 68.43, 50.99{]}
 & {[}60.61, 60.61, 60.61, 60.61, 60.61{]}
& {[}62.13, 65.52, 59.46, 55.89, 60.32{]}
& {[}54.55, 12.13, 96.97, 80, 21.06{]} & {[}61.21, 50.32, \textbf{\underline{72}}, 65.95, 52.31{]}
 \\[0.3cm]

\textbf{IFTSVM-L} & {[}65.16, 45.46, 84.85, 75, 56.61{]}
 & {[}65.16, 97.06, 31.25, 60, 74.16{]}
 & {[}63.64, 55.56, 69.24, 55.56, 55.56{]}
& {[}65.16, 55.89, 75, 70.38, 62.3{]}
 & {[}56.07, 22.59, 85.72, 58.34, 32.56{]} & {[}63.03, 55.31, 69.21, 63.85, 56.23{]}
 \\[0.3cm]

\textbf{IFTSVM-K} & {[}66.67, 60.61, 72.73, 68.97, 64.52{]}
 & {[}59.1, 21.88, 94.12, 77.78, 34.15{]}
 & {[}57.58, 86.21, 35.14, 51.03, 64.11{]}
 & {[}65.16, 65.72, 64.52, 67.65, 66.67{]}
 & {[}59.1, 46.15, 77.78, 75, 57.15{]} & {[}61.52, 56.11, 68.85, 68.08, 57.31{]}
 \\[0.3cm]

\textbf{LSSVM-L} & {[}63.64, 57.58, 69.7, 65.52, 61.3{]}
 & {[}60.61, 53.13, 67.65, 60.72, 56.67{]}
 & {[}57.58, 59.46, 55.18, 62.86, 61.12{]}
 & {[}63.64, 76.93, 55, 52.64, 62.5{]}
 & {[}60.61, 60, 61.12, 56.25, 58.07{]} & {[}61.21, 61.42, 61.73, 59.59, 59.93{]}
 \\[0.3cm]

\textbf{LSSVM-K} & {[}66.67, 68.58, 64.52, 68.58, 68.58{]}
 & {[}59.1, 66.67, 51.52, 57.9, 61.98{]}
 & {[}51.52, 64.71, 37.5, 52.39, 57.9{]}
 & {[}65.16, 84.85, 45.46, 60.87, 70.89{]}
 & {[}53.04, 71.43, 32.26, 54.35, 61.73{]} & {[}59.09, 71.24, 46.25, 58.81, 64.21{]}
 \\[0.3cm]

\textbf{LSTSVM-L} & {[}63.64, 69.7, 57.58, 62.17, 65.72{]}
 & {[}60.61, 29.42, 93.75, 83.34, 43.48{]}
 & {[}56.07, 33.34, 78.79, 61.12, 43.14{]}
& {[}62.13, 95.13, 8, 62.91, 75.73{]}
 & {[}59.1, 47.37, 75, 72, 57.15{]} & {[}60.3, 54.99, 62.62, 68.3, 57.04{]}
 \\[0.3cm]

\textbf{LSTSVM-K} & {[}57.58, 100, 0, 57.58, 73.08{]}
 & {[}62.13, 22.23, 89.75, 60, 32.44{]}
 & {[}60.61, 3.85, 97.5, 50, 7.15{]}
 & {[}60.61, 100, 18.75, 56.67, 72.35{]}
 & {[}54.55, 100, 0, 54.55, 70.59{]} & {[}59.09, 65.21, 41.2, 55.76, 51.12{]}
 \\[0.3cm]

\textbf{Linex-SVM-L} & {[}60.61, 53.34, 66.67, 57.15, 55.18{]}
 & {[}60.61, 58.83, 62.5, 62.5, 60.61{]}
 & {[}65.16, 58.63, 70.28, 60.72, 59.65{]}
 & {[}63.64, 55.18, 70.28, 59.26, 57.15{]}
 & {[}63.64, 73.08, 57.5, 52.78, 61.3{]} & {[}62.73, 59.81, 65.44, 58.48, 58.77{]}
 \\[0.3cm]

\textbf{Linex-SVM-K} & {[}59.1, 65.63, 52.95, 56.76, 60.87{]}
 & {[}59.1, 100, 0, 59.1, 74.29{]}
 & {[}60.61, 100, 0, 60.61, 75.48{]}
& {[}65.16, 100, 0, 65.16, 78.9{]}
     & {[}59.1, 100, 0, 59.1, 74.29{]} & {[}60.61, 93.13, 10.59, 60.14, \textbf{\underline{72.76}}{]}
 \\[0.3cm]

\textbf{Pin-SVM-L} & {[}62.13, 48.49, 75.76, 66.67, 56.15{]}
 & {[}59.1, 54.84, 62.86, 56.67, 55.74{]}
 & {[}56.07, 59.46, 51.73, 61.12, 60.28{]}
& {[}66.67, 59.46, 75.87, 75.87, 66.67{]}
 & {[}62.13, 83.88, 42.86, 56.53, 67.54{]} & {[}61.21, 61.22, 61.81, 63.37, 61.27{]}
\\[0.3cm]

\textbf{Pin-SVM-K} & {[}57.58, 100, 0, 57.58, 73.08{]}
 & {[}57.58, 100, 0, 57.58, 73.08{]}
 & {[}60.61, 100, 0, 60.61, 75.48{]}
 & {[}59.1, 67.57, 48.28, 62.5, 64.94{]}
 & {[}59.1, 100, 0, 59.1, 74.29{]} & {[}58.79, \textbf{\underline{93.51}}, 9.66, 59.47, 72.17{]}
\\[0.3cm]

\textbf{Pin-GTSVM-L} & {[}56.06,	63.33,	50,	51.35,	56.72{]}
 & {[}54.55,	57.14,	52.63,	47.06,	51.61{]}
 & {[}65.15,	67.86,	63.16,	57.58,	62.3{]}
 & {[}66.67,	65,	69.23,	76.47,	70.27{]}
 & {[}62.12,	77.42,	48.57,	57.14,	65.75{]} & {[}60.91, 66.15, 56.72, 57.92, 61.33{]}
\\[0.3cm]

\textbf{Pin-GTSVM-K} & {[}72.73,	73.33,	72.22,	68.75,	70.97{]}
 & {[}62.12,	65.52,	59.46,	55.88,	60.32{]}
 & {[}59.09,	57.58,	60.61,	59.36,	58.46{]}
 & {[}69.7,	86.67,	55.56,	61.9,	72.22{]}
 & {[}59.09,	57.14,	61.29,	62.5,	59.7{]} & {[}\textbf{\underline{64.55}}, 68.05, 61.83, 61.68, 64.33{]}
\\[0.3cm]
\hline
\textbf{Average} & {[}62.97, 70.71, 52.87, 62.93, 65.2{]} & {[}59.75, 61.2, \textbf{\underline{54.35}}, 61.56, 57.39{]} & {[}59.19, 62.43, 52.22, 57.77, 57.48{]} & {[}\textbf{\underline{64.3}}, \textbf{\underline{73.31}}, 52.5, \textbf{\underline{63.82}}, \textbf{\underline{67.12}}{]} & {[}59, 65.79, 50.19, 61.78, 59.61{]} \\[0.1cm]
\hline \\[-0.3cm]
\end{tabular}%
}
\footnotesize{The above performance is in \% and is mentioned in the format as follows: [Acc., Sens., Spec., Prec., f-measure].}
\footnotesize{The last column of the table is the average performance of each model for all the modalities. Boldface and underline in the last column refer to the best-performed model \textit{w.r.t.} each modality.}
\footnotesize{The last row of the table is the average performance contributed by each modality to the classification for all the models. Boldface and underline in the last row refer to the most prominent modality \textit{w.r.t.} Acc., Sens., Spec., Prec., and f-measure. The model $\mathcal{M}$ in the linear space is designated as $\mathcal{M}$-L, whereas its counterpart assessed in the Kernel space is referred to as $\mathcal{M}$-K.}
\end{table*}
% \usepackage{graphicx}
\begin{table}[]
\centering
\caption{Best parameters based on grid search technique for HbCs.}
\label{tab:Best_Parametrs_HbC}
\resizebox{\columnwidth}{!}{%
\begin{tabular}{|l|l|l|l|l|l|l|}
\hline
\multicolumn{1}{|c|}{\textbf{}} &
  \multicolumn{1}{c|}{\textbf{Parameters Sequence}} &
  \multicolumn{1}{c|}{\textbf{CT}} &
  \multicolumn{1}{c|}{\textbf{GM}} &
  \multicolumn{1}{c|}{\textbf{JD}} &
  \multicolumn{1}{c|}{\textbf{WM}} &
  \multicolumn{1}{c|}{\textbf{All Feature}} \\ \hline
\textbf{SVM-L} &
  ($C$) &
  (8) &
  (0.03125) &
  (0.03125) &
  (0.03125) &
  (0.03125) \\ \hline
\textbf{SVM-K} &
  ($C$, $\sigma$) &
  (2, 0.125) &
  (2, 64) &
  (0.5, 0.03125) &
  (0.5, 0.0625) &
  (0.5, 0.0625) \\ \hline
\textbf{TSVM-L} &
  ($C_1$, $C_2$) &
  (0.03125, 2) &
  (0.03125, 0.03125) &
  (0.03125, 0.03125) &
  (0.03125, 0.03125) &
  (0.03125, 0.03125) \\ \hline
\textbf{TSVM-K} &
  ($C_1$, $C_2$, $\sigma$) &
  (0.5, 2, 128) &
  (0.5, 0.125, 256) &
  (2, 0.5, 512) &
  (0.5, 0.5, 1024) &
  (0.125, 0.03125, 1024) \\ \hline
\textbf{IFTSVM-L} &
  ($C_1$, $C_2$, $\mu$) &
  (2, 2, 8) &
  (0.125, 32, 8) &
  (2, 0.5, 16) &
  (8, 2, 0.00097656) &
  (2, 0.03125, 32) \\ \hline
\textbf{IFTSVM-K} &
  ($C_1$, $C_2$, $\sigma$, $\mu$) &
  (2, 8, 128, 16) &
  (0.125, 0.03125, 64, 256) &
  (0.03125, 0.03125, 256, 128) &
  (2, 2, 1024, 64) &
  (0.5, 0.125, 512, 32) \\ \hline
\textbf{LSSVM-L} &
  ($C$) &
  (0.03125) &
  (0.125) &
  (0.03125) &
  (0.03125) &
  (0.03125) \\ \hline
\textbf{LSSVM-K} &
  ($C$, $\sigma$) &
  (32, 8) &
  (2, 32) &
  (0.5, 16) &
  (8, 16) &
  (0.5, 16) \\ \hline
\textbf{LSTSVM-L} &
  ($C_1$, $C_2$) &
  (0.03125, 0.5) &
  (0.125, 32) &
  (0.03125, 8) &
  (2, 0.5) &
  (2, 2) \\ \hline
\textbf{LSTSVM-K} &
  ($C_1$, $C_2$, $\sigma$) &
  (0.03125, 0.03125, 0.001953) &
  (0.5, 0.5, 256) &
  (0.03125, 0.125, 32) &
  (0.03125, 32, 32) &
  (32, 2, 128) \\ \hline
\textbf{Linex-SVM-L} &
  ($C$, $a$) &
  (0.03125, -8) &
  (0.125, -10) &
  (8, -5) &
  (0.125, -8) &
  (2, -2) \\ \hline
\textbf{Linex-SVM-L} &
  ($C$, $a$, $\sigma$) &
  (0.5, -4, 4) &
  (0.03125, -10, 0.000977) &
  (0.03125, -10, 0.000977) &
  (0.03125, -10, 0.000977) &
  (0.03125, -10, 0.000977) \\ \hline
\textbf{Pin-SVM-L} &
  ($C$, $\tau$) &
  (32, 0.2) &
  (0.03125, 0.6) &
  (0.125, 0.3) &
  (0.03125, 0.8) &
  (0.03125, 0) \\ \hline
\textbf{Pin-SVM-K} &
  ($C$, $\sigma$, $\tau$) &
  (0.125, 16, 0.7) &
  (8, 16, 0.1) &
  (0.125, 32, 0.5) &
  (0.5, 0.0625, 0) &
  (0.125, 64, 0.4) \\ \hline
\textbf{Pin-GTSVM-L} &
  ($C_1$, $C_2$, $\tau_1$, $\tau_2$) &
  (8, 8, 0.5, 0.5) &
  (0.03125, 0.03125, 0, 0) &
  (0.03125, 0.03125, 0.1, 0.1) &
  (0.03125, 0.03125, 0.3, 0.3) &
  (0.03125, 0.03125, 0.5, 0.5) \\ \hline
\textbf{Pin-GTSVM-K} &
  ($C_1$, $C_2$, $\sigma$, $\tau_1$, $\tau_2$) &
  (2, 0.5, 1024, 0.9, 0.9) &
  (0.03125, 0.03125, 128, 0.5, 0.5) &
  (2, 2, 1024, 0.8, 0.8) &
  (0.03125, 0.03125, 1024, 0.9, 0.9) &
  (0.03125, 0.03125, 512, 0, 0) \\ \hline
\end{tabular}%
}
\end{table}
% \usepackage{graphicx}
% \usepackage[normalem]{ulem}
% \usepackage{lscape}
% \begin{landscape}
\begin{table}[]
\centering
\caption{Rankings for HbCs with respect to accuracy.}
\label{tab:Rank_HbC}
\resizebox{\columnwidth}{!}{%
\begin{tabular}{|l|c|c|c|c|c|c|c|c|c|c|c|c|c|c|c|c|}
\hline
\multicolumn{1}{|c|}{\textbf{}} &
  \textbf{SVM-L} &
  \textbf{SVM-K} &
  \textbf{TSVM-L} &
  \textbf{TSVM-K} &
  \textbf{IFTSVM-L} &
  \textbf{IFTSVM-K} &
  \textbf{LSSVM-L} &
  \textbf{LSSVM-K} &
  \textbf{LSTSVM-L} &
  \textbf{LSTSVM-K} &
  \textbf{Linex-SVM-L} &
  \textbf{Linex-SVM-K} &
  \textbf{Pin-SVM-L} &
  \textbf{Pin-SVM-K} &
  \textbf{Pin-GTSVM-L} &
  \textbf{Pin-GTSVM-K} \\ \hline
\textbf{CT}          & 7.5 & 7.5  & 10.5 & 3    & 5    & 3    & 7.5  & 3    & 7.5  & 14.5 & 12  & 13  & 10.5 & 14.5 & 16  & 1                  \\ \hline
\textbf{GM}          & 14  & 14   & 10   & 3    & 1    & 10   & 6    & 10   & 6    & 3    & 6   & 10  & 10   & 14   & 16  & 3                  \\ \hline
\textbf{JD}          & 6.5 & 15.5 & 6.5  & 6.5  & 3    & 11.5 & 11.5 & 15.5 & 13.5 & 6.5  & 1.5 & 6.5 & 13.5 & 6.5  & 1.5 & 10                 \\ \hline
\textbf{WM}          & 1.5 & 14.5 & 10   & 12.5 & 6.5  & 6.5  & 10   & 6.5  & 12.5 & 14.5 & 10  & 6.5 & 3.5  & 16   & 3.5 & 1.5                \\ \hline
\textbf{All Feature} & 1.5 & 12.5 & 4    & 14.5 & 12.5 & 9    & 6    & 16   & 9    & 14.5 & 1.5 & 9   & 4    & 9    & 4   & 9                  \\ \hline
\textbf{Average}     & 6.2 & 12.8 & 8.2  & 7.9  & 5.6  & 8    & 8.2  & 10.2 & 9.7  & 10.6 & 6.2 & 9   & 8.3  & 12   & 8.2 & {\ul \textbf{4.9}} \\ \hline
\end{tabular}%
}
\end{table}
% \end{landscape}
%%%%%%%%%%%%%%%
The rank of each model concerning each modality is documented in Table \ref{tab:Rank_HbC}. The ultimate row in the table delineates the average rank assigned to each model. Impressively, the Pin-GTSVM-K secures the lowest average rank at $4.9$. This unequivocally indicates the superior performance of Pin-GTSVM-K relative to other HbCs. In the context of the Friedman test, with $\mathscr{N}=16$ and $\mathscr{P}=5$, we compute values of $\chi^2_F=16.01$ and $F_F=1.09$. From the statistical table for the $F$-distribution, we get $F_F (15, 60) = 1.84$ at a significance level of $5\%$. The null hypothesis can not be rejected due to the fact that $1.09 \ngeq 1.84$. Consequently, the Friedman test falls short in detecting statistical differences among the compared SOTA HbCs for the HC vs. SMC classification task. 

% \usepackage{graphicx}
% \usepackage{lscape}
% \begin{landscape}
\begin{table}[ht!]
\centering
\caption{Pairwise Win-tie-loss for HbCs.}
\label{tab:Win-Tie-Loss_HbC}
\resizebox{\columnwidth}{!}{%
\begin{tabular}{|l|c|c|c|c|c|c|c|c|c|c|c|c|c|c|c|}
\hline
 &
  SVM-L &
  SVM-K &
  TSVM-L &
  TSVM-K &
  IFTSVM-L &
  IFTSVM-K &
  LSSVM-L &
  LSSVM-K &
  LSTSVM-L &
  LSTSVM-K &
  Linex-SVM-L &
  Linex-SVM-K &
  Pin-SVM-L &
  Pin-SVM-K &
  Pin-GTSVM-L \\ \hline
SVM-K       & [$3, 2, 0$] &             &             &             &             &             &             &             &             &  &  &  &  &  &  \\ \hline
TSVM-L   & [$3, 1, 1$] & [$1, 0, 4$] &             &             &             &             &             &             &             &  &  &  &  &  &  \\ \hline
TSVM-K      & [$2, 1, 2$] & [$1, 0, 4$] & [$2, 1, 2$] &             &             &             &             &             &             &  &  &  &  &  &  \\ \hline
IFTSVM-L & [$2, 0, 3$] & [$0, 1, 4$] & [$1, 0, 4$] & [$1, 0, 4$] &             &             &             &             &             &  &  &  &  &  &  \\ \hline
IFTSVM-K    & [$3, 0, 2$] & [$0, 0, 5$] & [$2, 1, 2$] & [$2, 1, 2$] & [$2, 1, 2$] &             &             &             &             &  &  &  &  &  &  \\ \hline
LSSVM-L  & [$3, 1, 1$] & [$0, 1, 4$] & [$2, 1, 2$] & [$3, 0, 2$] & [$4, 0, 1$] & [$2, 1, 2$] &             &             &             &  &  &  &  &  &  \\ \hline
LSSVM-K     & [$3, 0, 2$] & [$1, 1, 3$] & [$2, 1, 2$] & [$3, 1, 1$] & [$3, 1, 1$] & [$2, 3, 0$] & [$3, 0, 2$] &             &             &  &  &  &  &  &  \\ \hline
LSTSVM-L & [$3, 1, 1$] & [$0, 1, 4$] & [$3, 0, 2$] & [$3, 1, 1$] & [$4, 0, 1$] & [$3, 1, 1$] & [$3, 2, 0$] & [$2, 0, 3$] &             &  &  &  &  &  &  \\ \hline
LSTSVM-K    & [$3, 1, 1$] & [$2, 1, 2$] & [$3, 1, 1$] & [$2, 3, 0$] & [$5, 0, 0$] & [$3, 0, 2$] & [$3, 0, 2$] & [$2, 0, 3$] & [$3, 0, 2$] &  &  &  &  &  &  \\ \hline
Linex-SVM-L &
  [$2, 1, 2$] &
  [$1, 0, 4$] &
  [$1, 1, 3$] &
  [$2, 0, 3$] &
  [$3, 0, 2$] &
  [$2, 0, 3$] &
  [$1, 2, 2$] &
  [$2, 0, 3$] &
  [$1, 1, 3$] &
  [$1, 0, 4$] &
   &
   &
   &
   &
   \\ \hline
Linex-SVM-K &
  [$3, 1, 1$] &
  [$1, 0, 4$] &
  [$2, 2, 1$] &
  [$2, 1, 2$] &
  [$3, 1, 1$] &
  [$1, 3, 1$] &
  [$3, 0, 2$] &
  [$1, 2, 2$] &
  [$2, 1, 2$] &
  [$1, 1, 3$] &
  [$4, 0, 1$] &
   &
   &
   &
   \\ \hline
Pin-SVM-L &
  [$4, 0, 1$] &
  [$1, 0, 4$] &
  [$1, 3, 1$] &
  [$3, 0, 2$] &
  [$3, 0, 2$] &
  [$2, 1, 2$] &
  [$3, 0, 2$] &
  [$1, 1, 3$] &
  [$2, 1, 2$] &
  [$2, 0, 3$] &
  [$3, 0, 2$] &
  [$1, 1, 3$] &
   &
   &
   \\ \hline
Pin-SVM-K &
  [$3, 2, 0$] &
  [$2, 1, 2$] &
  [$4, 1, 0$] &
  [$3, 1, 1$] &
  [$4, 0, 1$] &
  [$3, 1, 1$] &
  [$4, 0, 1$] &
  [$3, 0, 2$] &
  [$3, 1, 1$] &
  [$2, 2, 1$] &
  [$5, 0, 0$] &
  [$3, 2, 0$] &
  [$4, 0, 1$] &
   &
   \\ \hline
Pin-GTSVM-L &
  [$4, 0, 1$] &
  [$2, 0, 3$] &
  [$2, 1, 2$] &
  [$2, 0, 3$] &
  [$2, 0, 3$] &
  [$2, 0, 3$] &
  [$2, 0, 3$] &
  [$2, 0, 3$] &
  [$2, 0, 3$] &
  [$2, 0, 3$] &
  [$3, 1, 1$] &
  [$2, 0, 3$] &
  [$2, 2, 1$] &
  [$2, 0, 3$] &
   \\ \hline
Pin-GTSVM-K &
  [$2, 1, 2$] &
  [$0, 0, 5$] &
  [$2, 0, 3$] &
  [$1, 1, 3$] &
  [$2, 0, 3$] &
  [$0, 1, 4$] &
  [$1, 0, 4$] &
  [$0, 0, 5$] &
  [$0, 1, 4$] &
  [$1, 1, 3$] &
  [$2, 0, 3$] &
  [$1, 1, 3$] &
  [$1, 0, 4$] &
  [$1, 1, 3$] &
  [$2, 0, 3$] \\ \hline
  \multicolumn{9}{l}{In the $[\alpha,\beta,\gamma]$, $\alpha$, $\beta$, and $\gamma$ denote the number of win, tie, and loss of the row model over the corresponding column model.}\\
\end{tabular}%
}
\end{table}
% \end{landscape}
%%%%%%%%%%%%%%%%%%%%%%%%%

The pairwise Win-tie-loss results for Hyperplane-based Classifiers (HbCs) are presented in Table \ref{tab:Win-Tie-Loss_HbC}. The analysis reveals that SVM-K demonstrates statistical superiority over IFTSVM-K and Pin-GTSVM-K, LSTSVM-K outperforms IFTSVM-L, LSSVM-K surpasses Pin-GTSVM-K, and Pin-SVM-K prevails over Linex-SVM-L. Despite observing statistical differences between some HbCs, no single classifier emerges as dominant over others, in contrast to the findings in RNN-based classifiers as previously analyzed. 
%%%%%%%%%%%%%%%%%%%%%
Identifying crucial features of different modalities for the HC vs. SMC classification is vital. 
\subsection{Significant insights into the different features using Shapley}
\begin{figure}
\begin{minipage}{1\linewidth}
\centering
\subfloat[GM Modality]{\includegraphics[scale=0.28]{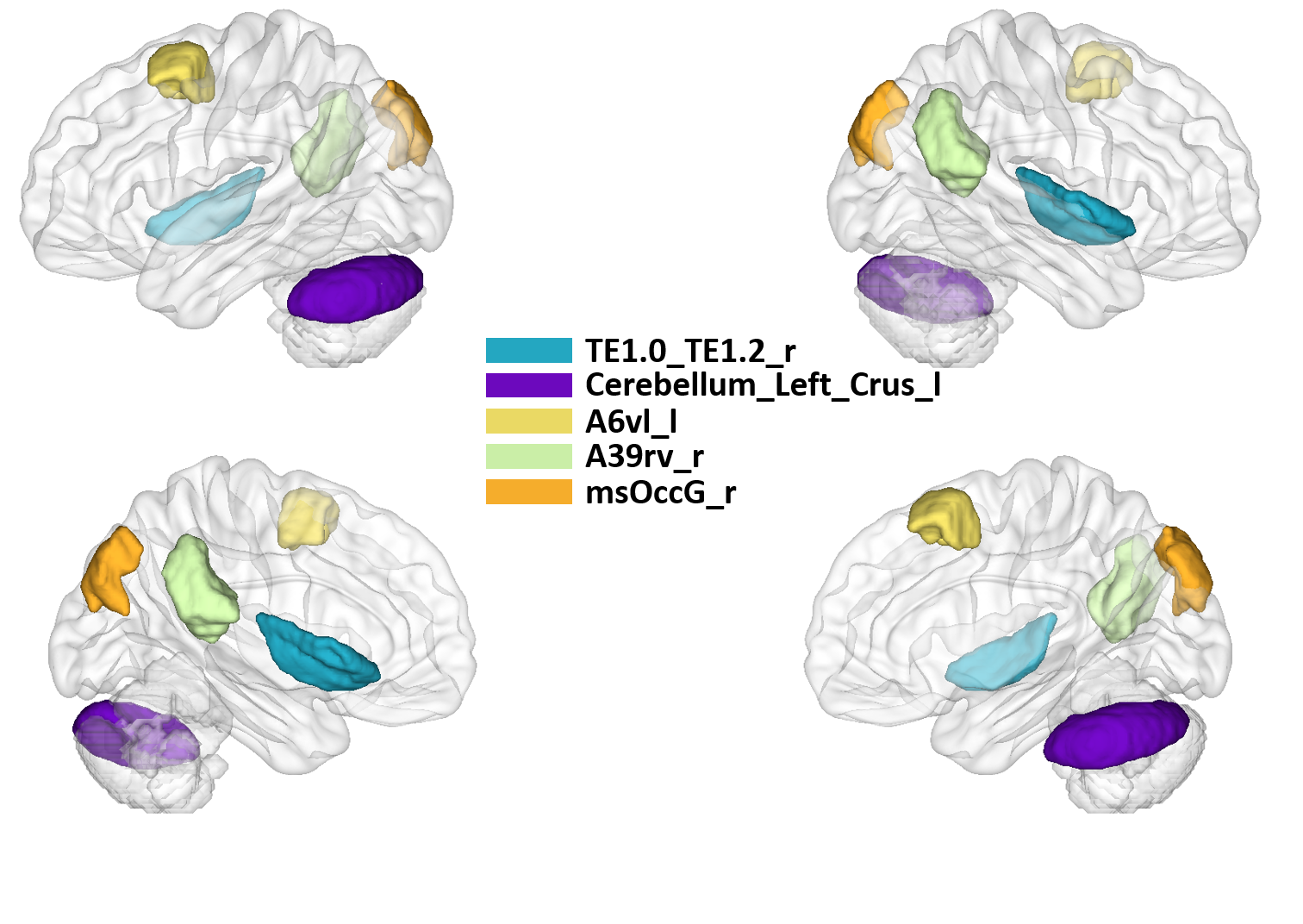}}
\end{minipage}
\par\medskip
% \par\medskip
\begin{minipage}{1\linewidth}
\centering
\subfloat[WM Modality]{\includegraphics[scale=0.28]{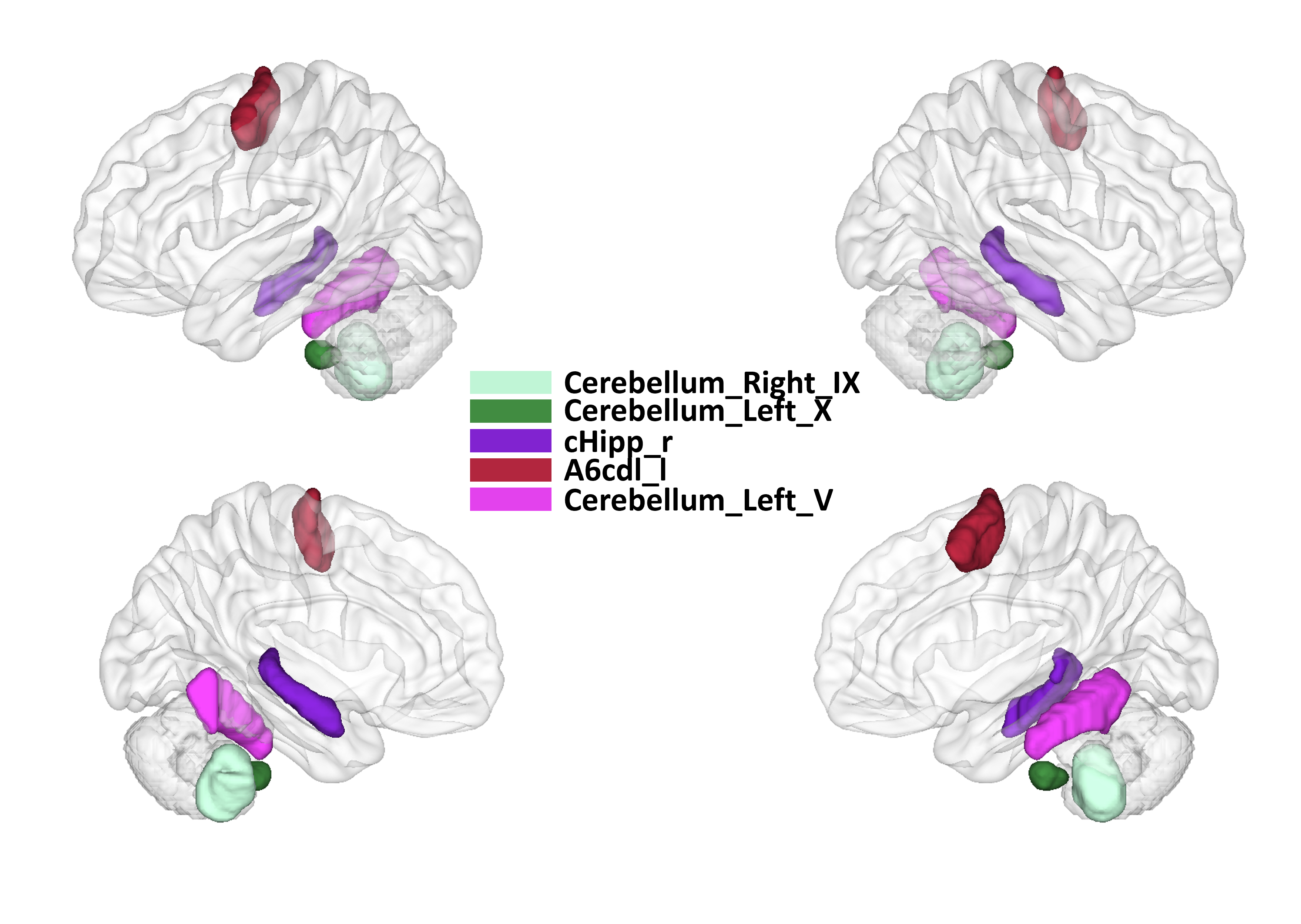}}
\end{minipage}
\par\medskip
% \par\medskip
\begin{minipage}{1\linewidth}
\centering
\subfloat[JD Modality]{\includegraphics[scale=0.28]{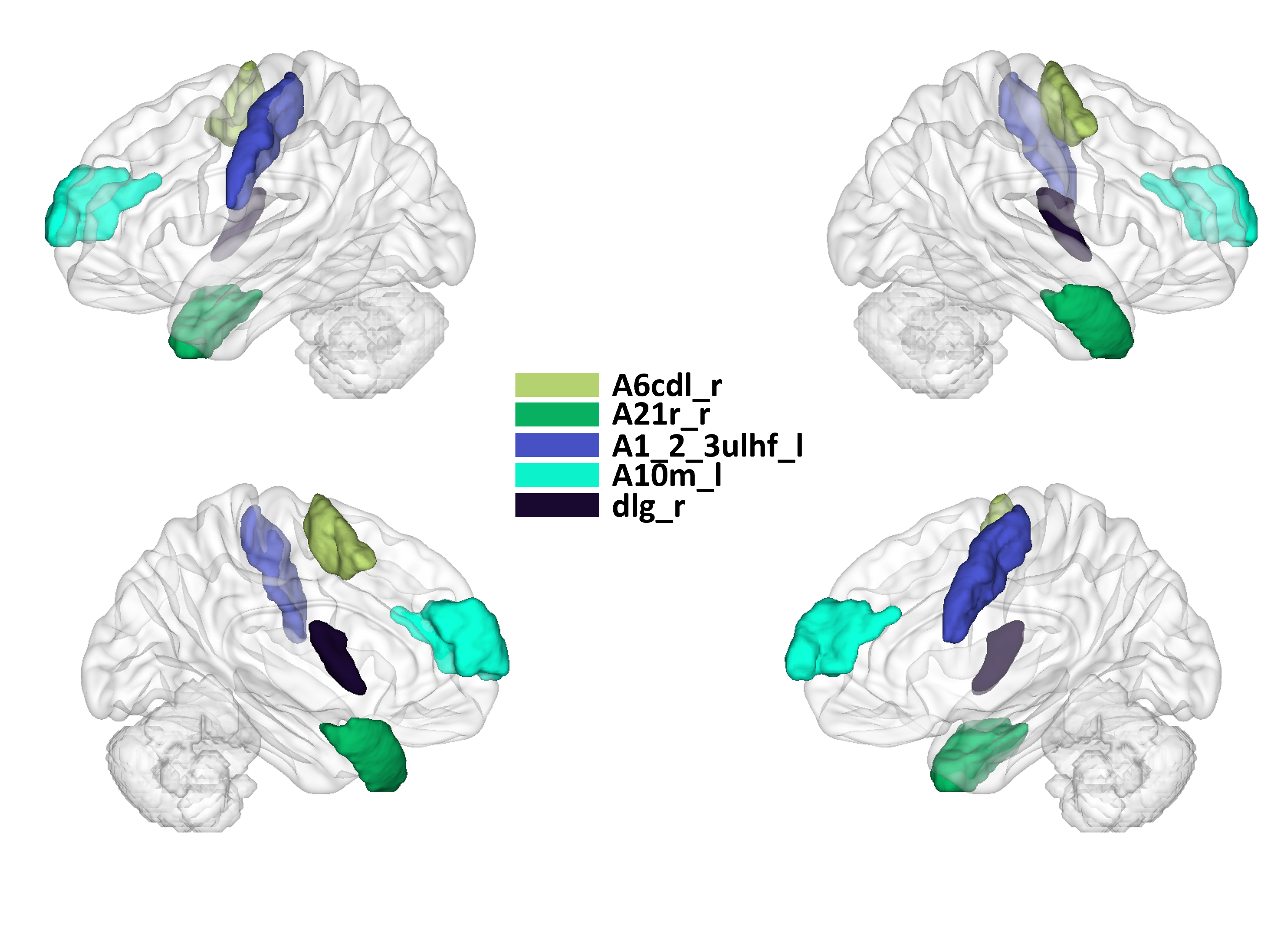}}
\end{minipage}
\par\medskip
\par\medskip
% \par\medskip
\begin{minipage}{1\linewidth}
\centering
\subfloat[CT Modality]{\includegraphics[scale=0.23]{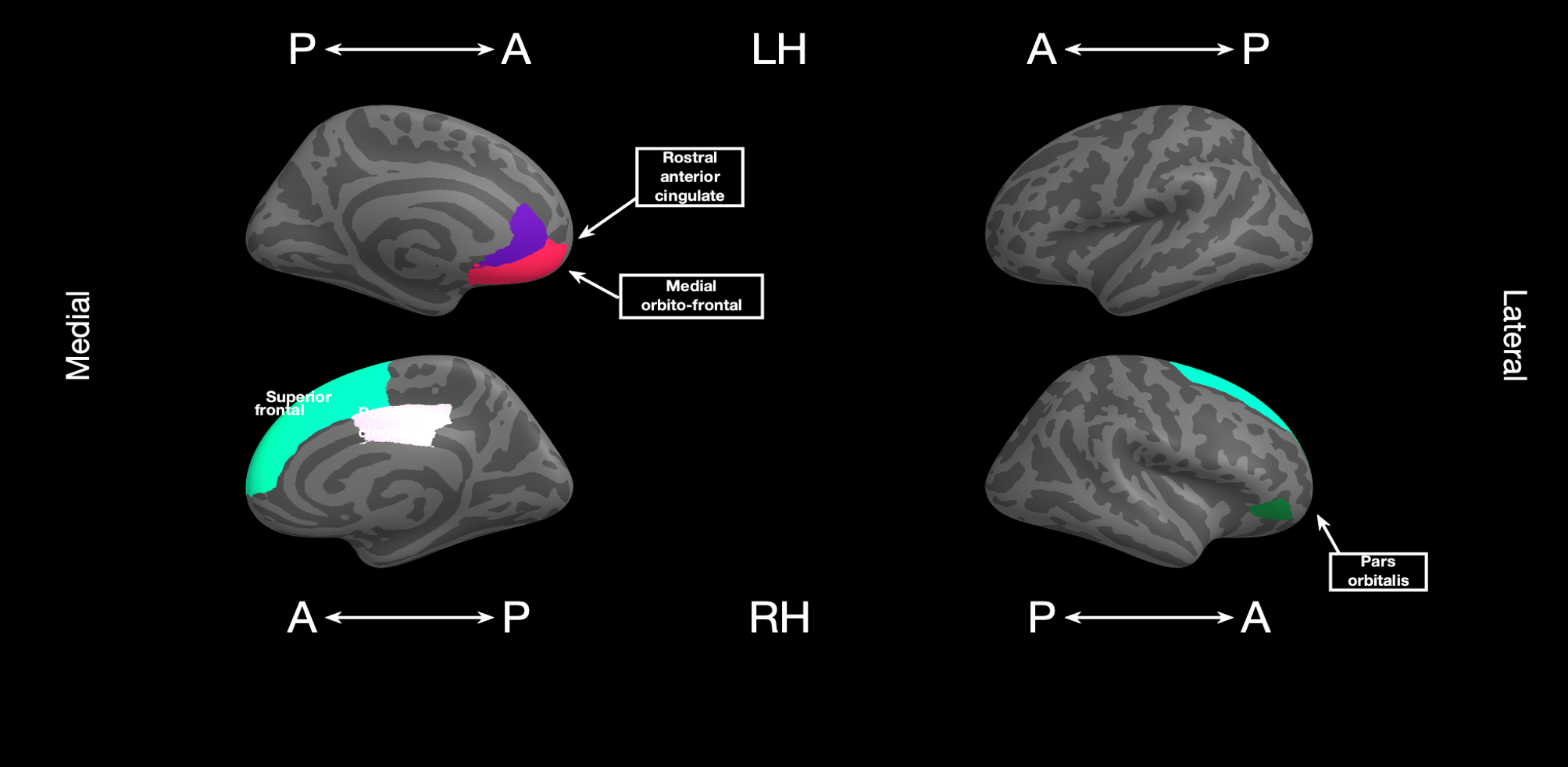}}
\end{minipage}
\par\medskip
\caption{Top five feature visualization for (a) GM, (b) WM, (c) JD, and (d) CT modalities identified from Shapley for dRVFL.}
\label{Fig:Shapley}
\end{figure}

Utilizing Shapley values \cite{molnar2020interpretable}, one of the most standard interpretability techniques in machine learning, we unveil insights into feature importance. Shapley values, rooted in cooperative game theory, justly allocate feature contributions, considering their interdependence. For this task, we selected the best-performing models, namely dRVFL from RNNs and Pin-GTSVM-K from HbCs. Over the best-performing models, we identified and visualized the top five features as shown in Figure \ref{Fig:Shapley}. 
These applied Shapley interpretable techniques shed light on the most pivotal features within the extracted array. This unveiling of crucial attributes ushers in a new era of comprehension, facilitating deeper insights into the intricate dynamics that underscore memory-related anomalies. The comprehensive discussion on feature interpretability is conducted in Section \ref{discussion}.
%%%%%%%%%%%
%%%%%%%%%%%%%%%%%%%%%%%%%%%%%%
\subsection{Discussion}
\label{discussion}
The study leveraged the ADNI dataset to thoroughly examine the binary classification task, discerning individuals with SMC from HC. The demographic profiles of both groups, with mean ages of 72.31 and 73.36 years, respectively, displayed no significant distinctions in terms of age or sex distribution, providing a solid foundation for comparison. Feature extraction encompassed 273 attributes across GM, WM, and JD metrics rooted in the Brainnetome atlas. Additionally, an extra 68 features pertaining to the cortical area were integrated, yielding a comprehensive and diverse information set. Applying bias correction techniques and segmentation into GM, WM, and cerebrospinal fluid scans effectively mitigated intensity variations, establishing a robust foundation for subsequent analysis.

In the subsequent subsections, we delve into the related findings concerning RNNs, HbCs models, feature interpretability, further key insights on SMC and Challenges in Utilizing Machine Learning for Identifying SMC and Possible Solutions.
%%%%%%%%%%
\begin{itemize}
\item \textbf{Insights from RNN Models:} Beyond feature extraction, this study extended its ambit to evaluate diverse classification models. The careful selection of RNNs reflects a nuanced consideration of model complexity and aggregation. These models exhibited commendable performance across distinct feature sets, with dRVFL excelling notably in four sets (GM, WM, JD, and the combined feature set), while edRVFL showcased superior performance in the CT feature set. The results bear significant import for the field of cognitive health assessment. The proficiency demonstrated by dRVFL underscores the salience of deeper, multi-layered architectures in capturing intricate cognitive patterns.On the other hand, the success of edRVFL in the CT feature set implies that ensemble methods can perform very well in situations where individual anatomical intricacies play a crucial role. These findings resonate with prior research, underlining the import of judicious model selection in concert with the nature of extracted features.

\item \textbf{Insights from HbC Models:} Through a rigorous evaluation encompassing an array of 16 distinct SVM variants, we meticulously assessed their performance across feature sets extracted from diverse modalities. Notably, for attributes related to CT and WM, Pin-GTSVM-K emerged as the most proficient performer, showcasing superior accuracy. Similarly, for the JD feature set, Pin-GTSVM-L demonstrated highly commendable performance. In contrast, for GM attributes, IFTSVM-L emerged as the standout performer. When amalgamating all features, Linex-SVM and SVM-L proved to be the two most efficient models, exhibiting superior performance metrics. These findings carry substantial significance for the field of cognitive health assessment. The subtle disparities in performance among SVM variants underscore the pivotal role of judicious model selection. The demonstrated proficiency of Pin-GTSVM-K in managing CT and WM attributes implies its appropriateness in situations where complex anatomical intricacies are influential.
Similarly, the robust performance of Pin-GTSVM-L in the JD feature set indicates its proficiency in capturing nuanced cognitive intricacies. The ascendancy of IFTSVM-L as the leading performer for GM attributes suggests its effectiveness in contexts where GM characteristics play a decisive role.

\item \textbf{Feature Interpretability:} The highest prediction accuracy of 78.79\% was achieved by combining all features (GM, WM, JD, and CT) using the dRVFL algorithm (with SEN = 84.62\% and SPE = 75.78\%; Table 5). In this study, the Shapely feature importance technique was used to determine the most significant features in each model \cite{chan2023alzheimer}. Since the dRVFL algorithm achieved the best prediction performance, our focus was solely on conducting Shapley analysis through this algorithm.
In the GM metric, notable findings were observed in the primary auditory cortex, cerebellum left crus, ventrolateral area 6, rostroventral area 39, and medial superior occipital gyrus regions (Table \ref{featuretable}). Even though the primary auditory cortex is mainly associated with processing auditory information, it also plays a role in emotion and anxiety, which are preclinical features for Alzheimer's disease \cite{micheyl2007role, swords2018auditory}. The cerebellum was also found to be an important GM region in our analysis. The cerebellum is mostly known for coordination and movement control, but it is also responsible for cognition, emotions, and decision-making \cite{jacobs2018cerebellum}. The SHAP analysis also identified the ventrolateral area 6, rostroventral area 39, and medial superior occipital gyrus regions as the most significant, all of which are directly linked to attention, cognition, and social cognition.
In the WM modality, interestingly, the SHAP analysis revealed the most important regions in the cerebellum and hippocampus regions (Table \ref{featuretable}). The hippocampus, a well-known region in the limbic system responsible for memory and cognition, is a crucial area in studies of Alzheimer's disease \cite{rao2022hippocampus}.  This analysis, along with the results from the GM analysis, suggests that cerebellum features could serve as promising biomarkers for identifying AD subjects before clinical symptoms appear. While some studies have highlighted the cerebellum's role in the development, cognitive deficits, and dysconnectivity linked to AD \cite{jacobs2018cerebellum,potvin2019association}, our understanding of its involvement remains limited, necessitating further research.
In addition to our primary focus, we delved into the JD features for SMC identification. It is crucial to note that JD values derived from Deformation-Based Morphometry (DBM) are quantitative measures used to assess the extent of local tissue expansion or compression in brain images. Elevated JD values indicate tissue expansion, while lower JD values suggest tissue compression. Studies have demonstrated that DBM is a reliable approach for detecting brain atrophy in subcortical regions \cite{tremblay2020sex}.  The dRVFL algorithm applied to JD features resulted in a commendable accuracy of 74.25\%, with a sensitivity of 81.49\% and a specificity of 69.24\% (Table \ref{rnntab}). Notably, SMC and HC subjects have similar clinical measurements (insignificant), and clinically, the SMC subjects are not classified as MCI or AD. As a result, the classification between these two groups would be one of the challenging tasks in the area of AD. In the SHAP analysis on JD features, it was found that the top regions were mainly located in the Frontal Lobe, and they are primarily associated with Imagination, cognition, memory, and language [Table \ref{featuretable}].
Cortical thinning is considered to be one of the most important and informative features in AD \cite{wu2021cortical}. Our SHAP analysis revealed significant regions in medial orbitofrontal cortex, frontal pole, superior parietal, rostral middle frontal, and fusiform gyrus. Interestingly, these regions are mainly associated with cognitive processes such as decision-making, social cognition, working memory, cognitive control, and face and object recognition.
%%%%%%%%%%%%%%%%%%
\item \textbf{Further Key Insights on SMC and its Implications:} SMC represents a cognitive status between cognitively healthy individuals and those suffering from MCI. In the ADNI dataset, attempts have been made to gather data on this group from the ADNI-2 phase to establish a link between cognitive health and MCI. Individuals with SMC may experience subjective memory issues or changes that are seldom noticeable through conventional cognitive tests, yet they do not significantly impact their daily lives \cite{hsieh2019effects}. It has been documented that SMC might be an early sign of accelerated decline in cognitive functioning, which may manifest later in this population \cite{liew2020subjective,yates2017subjective}. Furthermore, individuals with SMC have been found to have a higher likelihood of developing AD. Hence, it is suggested that SMC be regarded as an additional risk factor for the onset of AD, and potentially as a pre-MCI state \cite{warren2022subjective}.  According to a meta-analysis \cite{mitchell2008clinical}, individuals with dementia have a SMC status of 42.8\%, while those with MCI have a rate of 38.2\%, in comparison to a rate of 17.4\% in cognitively healthy subjects. Therefore, detecting this cohort of individuals in the initial phases of cognitive impairment would greatly aid in administering preventative care for them \cite{miley2019lifestyle}. One interesting aspect of SMC treatment is that individuals with SMC may not need AD chemical medications, such as Aducanumab, which can have significant side effects. Instead, it is hypothesized that they could benefit from psychological interventions to help manage SMC by reducing stress, improving sleep, and changing their lifestyle \cite{warren2022subjective}. This strategy has the potential to decrease the costs linked to unnecessary procedures, alleviate the strain on cognitive clinics, and result in symptom improvement irrespective of cognitive impairment \cite{warren2022subjective}. Overall, these points emphasize the importance of early detection and differentiation of SMC subjects from cognitively healthy individuals in clinical settings, which is the main focus of this study.

It is crucial to emphasize that the SMC data in ADNI has been obtained through a self-reported procedure, where a pivotal inclusion criterion involves participants self-reporting significant memory concerns. Self-reported data may be susceptible to self-reporting bias. Indeed, the precision and dependability of self-reported information may be influenced by factors such as mood, fatigue, and external pressures. Consequently, our findings should be interpreted with caution, and further studies with larger sample sizes are necessary to validate our results. To the best of our knowledge, this study is one of the pioneering efforts to utilize ML models for identifying SMC subjects from healthy controls. Our future plans include the use of deep learning models in this area. 
%%%%%%%%%%%%%%%%%%%%%%%%%%
\item \textbf{Challenges in Utilizing Machine Learning for Identifying SMC and Their Possible Solutions:} Certainly, discussing potential biases, overfitting issues, and scalability challenges in clinical settings is imperative for refining machine learning models and guiding future research efforts effectively. Biases, such as sample selection bias or data imbalance, can significantly impact the generalizability of models, potentially leading to skewed results or inaccurate predictions. By acknowledging and addressing these biases, researchers can ensure that their models are representative of diverse populations and applicable beyond specific contexts.
Overfitting is another critical concern that researchers must confront and occurs when a model learns noise in the data rather than the underlying patterns, resulting in poor performance when applied to new data. To mitigate overfitting, rigorous validation techniques and regularization methods are essential \cite{hastie2009elements}. These techniques help ensure that models capture meaningful patterns in the data without being overly influenced by noise or outliers.

Moreover, scalability \cite{gesi2023leveraging} is a crucial consideration, especially concerning the practical implementation of machine learning models in clinical settings. Factors such as computational resources, data privacy regulations, and interpretability with existing healthcare systems can significantly impact the scalability of these models. Researchers must carefully assess these factors and develop strategies to address scalability challenges to ensure the widespread adoption and utility of their models in real-world clinical environments. Introducing fuzzy logic in machine learning models generally enhances the interpretability of the model \cite{2024sajidnfrvfl}, and one can explore the fusion of fuzzy logic with state-of-the-art models to make it interpretable and more relevant in biomedical domains such as AD diagnosis. 
By proactively discussing and addressing biases, overfitting issues, and scalability challenges, researchers can enhance the robustness and reliability of machine learning models for clinical applications. This comprehensive approach not only improves the accuracy and effectiveness of these models but also fosters trust and confidence among clinicians and healthcare professionals in leveraging machine learning technology to improve patient care and outcomes.
\end{itemize}

% \usepackage{graphicx}
\begin{table}[]
\centering
\caption{Top five significant features identified through SHAP analysis across different anatomical MRI matrices using the dRVFL algorithm.}
\label{featuretable}
\resizebox{\textwidth}{!}{%
\begin{tabular}{|c|l|l|l|l|}
\hline
\textbf{Top 5 Features} & \multicolumn{1}{c|}{\textbf{GM}} & \multicolumn{1}{c|}{\textbf{WM}} & \multicolumn{1}{c|}{\textbf{JD}} & \multicolumn{1}{c|}{\textbf{CT}} \\ \hline
\textbf{1} & TE1.0 and TE1.2 (R) & Cerebellum IX (R) & Caudal dorsolateral area 6 (L) & Medial orbitofrontal cortex (L) \\ \hline
\textbf{2} & Cerebellum Left Crus (L) & Cerebellum X (L) & Rostral area 21 (R) & Frontal pole (L) \\ \hline
\textbf{3} & Ventrolateral area 6 (L) & Caudal hippocampus (R) & Area 1/2/3 (upper limb, head and face   region) (L) & Superior parietal (R) \\ \hline
\textbf{4} & Rostroventral area 39(R) & Caudal dorsolateral area 6 (L) & Medial area 10 (L) & Rostral middle frontal (R) \\ \hline
\textbf{5} & Medial superior occipital gyrus (R) & Cerebellum V (L) & Dorsal granular insular (R) & Fusiform Gyrus (R) \\ \hline  
\end{tabular}}
\footnotesize{Note: CT: Cortical Thickness; GM: Gray Matter; JD: Jacobian determinant; WM: White Matter. The letter inside the parentheses stands for right (R) or left (L) hemispheres. The Brainnetome atlas was used as the source for the brain regions \cite{fan2016human}, and DKT atlas was the source  for CT regions \cite{klein2012101}.}
\end{table}

\section{Conclusion and Future Scope}
\label{conclusion}
This study undertakes an exhaustive review of the current literature, including cutting-edge reviews, survey papers, and pertinent contributing articles. It underscores the importance of conducting a comprehensive assessment of SMC identification. Early and accurate cognitive health diagnosis is essential for targeted intervention and individualized therapy. Through extensive analysis, we have brought to light the intricate subtleties that underlie the manifestation of SMC, offering insight into its escalating impact as populations age. The deliberate selection of two distinct methodologies, RNNs, and HbCs, was driven by strategic intent. Each approach brings forth unique strengths. RNNs adeptly unravel complex, non-linear relationships within the data, providing a nuanced perspective on cognitive health patterns. HbCs leverage the concept of hyperplanes for precise boundary definition, presenting an alternative yet equally valuable framework for discerning cognitive disparities.

This study not only adds depth to the expanding field of cognitive health assessment but also underscores the importance of methodological diversity in research endeavors of this nature. Through a multi-dimensional evaluation encompassing diverse models and feature sets, we endeavor to establish a strong foundation for future explorations in this critical field. Furthermore, the assessment of feature importance carried out through advanced techniques, such as Shapley values, furnished invaluable insights into the discriminative capacity of individual features. Additionally, comprehensive statistical tests are meticulously employed to validate the models' statistical significance and credibility.

Future research seeks, while the current study remains independent of the most recent advancements in neuroimaging and machine learning, staying attuned to emerging technologies is imperative. The seamless integration of any forthcoming breakthroughs in these domains can significantly refine the precision of discerning SMCs from healthy control (HC) subjects in forthcoming investigations. Expanding the horizons beyond SMC, the profound insights gleaned from this exhaustive assessment of RNNs and HbCs can be extrapolated to a wider spectrum of cognitive health considerations. The methodologies delineated here form a promising platform for delving into various other neurological and psychiatric conditions.

Future research endeavors could delve into the synergistic interplay between diverse imaging modalities or supplementary cognitive assessments, thereby fine-tuning the identification and characterization of SMCs. The validation and eventual deployment of HbCs in clinical settings signify a propitious trajectory. Emphasizing preprocessing techniques is pivotal, ensuring that the models exhibit robustness and reliability in real-world applications. This undertaking would necessitate a systematic validation process encompassing clinical data and expert evaluation. In light of potential progressions in the field, exploring hybrid models or ensemble techniques is an intriguing avenue for future research. By harnessing the collective strengths of disparate models and feature sets, the prospect of achieving an even more precise and dependable identification of SMC beckons. Further incursions into this domain of exploration and experimentation promise to be profoundly beneficial. We trust that this work will serve as a beacon, igniting further research and innovation in pursuing elevated cognitive health and well-being. The source codes of the algorithms and datasets used in this study are available at
\emph{https://github.com/mtanveer1/SMC}.

{\section*{ACKNOWLEDGEMENT}
This study is supported by the Indian government’s Department of Science and Technology (DST) in collaboration with the Ministry of Electronics and Information Technology (MeITy) Grant No. $DST/NSM/R\&D\_HPC\_Appl/2021/03.29$ and Science and Engineering Research Board under the Mathematical Research Impact-Centric Support (MATRICS) scheme Grant No. MTR/2021/000787. M. Sajid also acknowledges support from the Council of Scientific and Industrial Research (CSIR), New Delhi for providing fellowship under the grants 09/1022(13847)/2022-EMR-I. The dataset employed in this study was procured with the aid of financial support from the Alzheimer's Disease Neuroimaging Initiative (ADNI), which was made possible through the National Institutes of Health's U01 AG024904 grant and the Department of Defense's ADNI award W81XWH-12-2-0012. The aforementioned initiative's funding was sourced from the National Institute on Aging, the National Institute of Biomedical Imaging and Bioengineering, and several munificent contributions made by a variety of entities: F. Hoffmann-La Roche Ltd. and its affiliated company Genentech, Inc.; Bristol-Myers Squibb Company;  Alzheimer’s Drug Discovery Foundation; Merck \& Co., Inc.;  CereSpir, Inc.; Meso Scale Diagnostics, LLC.;  Novartis Pharmaceuticals Corporation; AbbVie, Alzheimer’s Association; Lumosity; Biogen; Fujirebio; IXICO Ltd.; Araclon Biotech; BioClinica, Inc.; NeuroRx Research; EuroImmun; Piramal Imaging; GE Healthcare; Cogstate; Eisai Inc.; Johnson \& Johnson Pharmaceutical Research \& Development LLC.;  Servier; Eli Lilly and Company; Transition Therapeutics Elan Pharmaceuticals, Inc.; Janssen Alzheimer Immunotherapy Research \& Development, LLC.; Lundbeck; Neurotrack Technologies; Pfizer Inc. and Takeda Pharmaceutical Company. Financial aid from the Canadian Institutes of Health Research is being extended to sustain ADNI clinical sites across Canada. Meanwhile, the Foundation for the National Institutes of Health (www.fnih.org) has facilitated private sector donations to support this endeavor.  Support for the grants dedicated to research and education was furnished by the Northern California Institute and the Alzheimer's Therapeutic Research Institute at the University of Southern California. The data associated with the ADNI initiative were made available through the auspices of the Neuro Imaging Laboratory located at the University of Southern California. The present study relied upon the ADNI dataset, which can be accessed via adni.loni.usc.edu. The ADNI initiative was planned and executed by the ADNI investigators, although they did not contribute to either the analysis or writing of this particular article. A thoroughly detailed listing of ADNI investigators can be accessed via the following link: \url{http://adni.loni.usc.edu/wp-content/uploads/how_to_apply/ADNI_Acknowledgment_List.pdf}.

\section*{CONFLICT OF INTEREST STATEMENT}
All authors declare no conflicts of interest.

 \bibliography{AARef}

\begin{thebibliography}{80}
\expandafter\ifx\csname natexlab\endcsname\relax\def\natexlab#1{#1}\fi
\expandafter\ifx\csname url\endcsname\relax
  \def\url#1{{\tt #1}}\fi
\expandafter\ifx\csname urlprefix\endcsname\relax\def\urlprefix{URL }\fi

\bibitem[{Adelson et~al.(2023)Adelson, Garikipati, Maharjan, Ciobanu, Barnes,
  Singh, Dinenno, Mao, \& Das}]{adelson2023machine}
Adelson, R.~P., Garikipati, A., Maharjan, J., Ciobanu, M., Barnes, G., Singh,
  N.~P., Dinenno, F.~A., Mao, Q., \& Das, R. (2023).
\newblock Machine learning approach for improved longitudinal prediction of
  progression from mild cognitive impairment to {A}lzheimer’s {D}isease.
\newblock {\em Diagnostics\/}, {\em 14\/}(1), 13.

\bibitem[{Altwijri et~al.(2023)Altwijri, Alanazi, Aleid, Alhussaini, Aloqalaa,
  Almijalli, \& Saad}]{altwijri2023novel}
Altwijri, O., Alanazi, R., Aleid, A., Alhussaini, K., Aloqalaa, Z., Almijalli,
  M., \& Saad, A. (2023).
\newblock Novel deep-learning approach for automatic diagnosis of alzheimer’s
  disease from {MRI}.
\newblock {\em Applied Sciences\/}, {\em 13\/}(24), 13051.

\bibitem[{Arya et~al.(2023)Arya, Verma, Chakarabarti, Chakrabarti, Elngar,
  Kamali, \& Nami}]{Arya2023}
Arya, A.~D., Verma, S.~S., Chakarabarti, P., Chakrabarti, T., Elngar, A.~A.,
  Kamali, A.~M., \& Nami, M. (2023).
\newblock A systematic review on machine learning and deep learning techniques
  in the effective diagnosis of alzheimer’s disease.
\newblock {\em Brain Informatics\/}, {\em 10\/}.

\bibitem[{{\'A}vila-Jim{\'e}nez et~al.(2023){\'A}vila-Jim{\'e}nez,
  Cant{\'o}n-Habas, del Pilar Carrera-Gonz{\'a}lez, Rich-Ruiz, \&
  Ventura}]{avila2023deep}
{\'A}vila-Jim{\'e}nez, J., Cant{\'o}n-Habas, V., del Pilar
  Carrera-Gonz{\'a}lez, M., Rich-Ruiz, M., \& Ventura, S. (2023).
\newblock A deep learning model for {A}lzheimer’s {D}isease diagnosis based
  on patient clinical records.
\newblock {\em Computers in Biology and Medicine\/}, (p. 107814).

\bibitem[{Bento et~al.(2022)Bento, Fantini, Park, Rittner, \&
  Frayne}]{Bento2022}
Bento, M., Fantini, I., Park, J., Rittner, L., \& Frayne, R. (2022).
\newblock Deep learning in large and multi-site structural brain mr imaging
  datasets.
\newblock {\em Frontiers in Neuroinformatics\/}, {\em 15\/}.

\bibitem[{Chan et~al.(2023)Chan, Fischer, Maralani, Black, Moody, \&
  Khademi}]{chan2023alzheimer}
Chan, K., Fischer, C., Maralani, P.~J., Black, S.~E., Moody, A.~R., \& Khademi,
  A. (2023).
\newblock Alzheimer’s and vascular disease classification using regional
  texture biomarkers in flair mri.
\newblock {\em NeuroImage: Clinical\/}, {\em 38\/}, 103385.

\bibitem[{Chen \& Liu(2018)}]{7987745}
Chen, C. L.~P., \& Liu, Z. (2018).
\newblock Broad learning system: An effective and efficient incremental
  learning system without the need for deep architecture.
\newblock {\em IEEE Transactions on Neural Networks and Learning Systems\/},
  {\em 29\/}(1), 10--24.

\bibitem[{Cortes \& Vapnik(1995)}]{cortes1995support}
Cortes, C., \& Vapnik, V. (1995).
\newblock Support-vector networks.
\newblock {\em Machine Learning\/}, {\em 20\/}, 273--297.

\bibitem[{Dem{\v{s}}ar(2006)}]{demvsar2006statistical}
Dem{\v{s}}ar, J. (2006).
\newblock Statistical comparisons of classifiers over multiple data sets.
\newblock {\em The Journal of Machine Learning Research\/}, {\em 7\/}, 1--30.

\bibitem[{Fan et~al.(2023)Fan, Yang, Zhang, Peng, Zhou, Liu, Chen, \&
  Hou}]{10332232}
Fan, C.-C., Yang, H., Zhang, C., Peng, L., Zhou, X., Liu, S., Chen, S., \& Hou,
  Z.-G. (2023).
\newblock Graph reasoning module for {A}lzheimer’s {D}isease diagnosis: A
  plug-and-play method.
\newblock {\em IEEE Transactions on Neural Systems and Rehabilitation
  Engineering\/}, {\em 31\/}, 4773--4780.

\bibitem[{Fan et~al.(2016)Fan, Li, Zhuo, Zhang, Wang, Chen, Yang, Chu, Xie,
  Laird et~al.}]{fan2016human}
Fan, L., Li, H., Zhuo, J., Zhang, Y., Wang, J., Chen, L., Yang, Z., Chu, C.,
  Xie, S., Laird, A.~R., et~al. (2016).
\newblock The human brainnetome atlas: a new brain atlas based on connectional
  architecture.
\newblock {\em Cerebral Cortex\/}, {\em 26\/}(8), 3508--3526.

\bibitem[{Farokhian et~al.(2017)Farokhian, Beheshti, Sone, \&
  Matsuda}]{farokhian2017comparing}
Farokhian, F., Beheshti, I., Sone, D., \& Matsuda, H. (2017).
\newblock Comparing cat12 and vbm8 for detecting brain morphological
  abnormalities in temporal lobe epilepsy.
\newblock {\em Frontiers in Neurology\/}, {\em 8\/}, 428.

\bibitem[{Fathi et~al.(2022)Fathi, Ahmadi, \& Dehnad}]{Fathi2022}
Fathi, S., Ahmadi, M., \& Dehnad, A. (2022).
\newblock Early diagnosis of alzheimer's disease based on deep learning: A
  systematic review.
\newblock {\em Computers in biology and medicine\/}, {\em 146\/}, 105634.

\bibitem[{Feng \& Chen(2020)}]{8432091}
Feng, S., \& Chen, C.~P. (2020).
\newblock Fuzzy broad learning system: A novel neuro-fuzzy model for regression
  and classification.
\newblock {\em IEEE Transactions on Cybernetics\/}, {\em 50\/}(2), 414--424.

\bibitem[{Filley(1998)}]{filley1998behavioral}
Filley, C.~M. (1998).
\newblock The behavioral neurology of cerebral white matter.
\newblock {\em Neurology\/}, {\em 50\/}(6), 1535--1540.

\bibitem[{Ganaie et~al.(2024,
  https://doi.org/10.1109/TNNLS.2024.3353531)Ganaie, Sajid, Malik, \&
  Tanveer}]{ganaie2023graph}
Ganaie, M., Sajid, M., Malik, A., \& Tanveer, M. (2024,
  https://doi.org/10.1109/TNNLS.2024.3353531).
\newblock Graph embedded intuitionistic fuzzy random vector functional link
  neural network for class imbalance learning.
\newblock {\em IEEE Transactions on Neural Networks and Learning Systems\/}.

\bibitem[{Ganaie et~al.(2022)Ganaie, Hu, Malik, Tanveer, \&
  Suganthan}]{ganaie2022ensemble}
Ganaie, M.~A., Hu, M., Malik, A., Tanveer, M., \& Suganthan, P. (2022).
\newblock Ensemble deep learning: A review.
\newblock {\em Engineering Applications of Artificial Intelligence\/}, {\em
  115\/}, 105151.

\bibitem[{Ganaie et~al.(2020)Ganaie, Tanveer, \& Suganthan}]{ganaie2020minimum}
Ganaie, M.~A., Tanveer, M., \& Suganthan, P.~N. (2020).
\newblock Minimum variance embedded random vector functional link network.
\newblock (pp. 412--419). \textit{Neural Information Processing: 27th
  International Conference, ICONIP 2020, Bangkok, Thailand, November 18--22,
  Springer, Proceedings, Part V 27}.

\bibitem[{Gauthier et~al.(2006)Gauthier, Reisberg, Zaudig, Petersen, Ritchie,
  Broich, Belleville, Brodaty, Bennett, Chertkow et~al.}]{gauthier2006mild}
Gauthier, S., Reisberg, B., Zaudig, M., Petersen, R.~C., Ritchie, K., Broich,
  K., Belleville, S., Brodaty, H., Bennett, D., Chertkow, H., et~al. (2006).
\newblock Mild cognitive impairment.
\newblock {\em The Lancet\/}, {\em 367\/}(9518), 1262--1270.

\bibitem[{Gesi et~al.(2023)Gesi, Shen, Geng, Chen, \&
  Ahmed}]{gesi2023leveraging}
Gesi, J., Shen, X., Geng, Y., Chen, Q., \& Ahmed, I. (2023).
\newblock Leveraging feature bias for scalable misprediction explanation of
  machine learning models.
\newblock In {\em 2023 IEEE/ACM 45th International Conference on Software
  Engineering (ICSE)\/}, (pp. 1559--1570). \textit{2023 IEEE/ACM 45th
  International Conference on Software Engineering (ICSE)}.

\bibitem[{Goel et~al.(2023)Goel, Sharma, Tanveer, Suganthan, Maji, \&
  Pilli}]{goel2023multimodal}
Goel, T., Sharma, R., Tanveer, M., Suganthan, P.~N., Maji, K., \& Pilli, R.
  (2023).
\newblock Multimodal neuroimaging based alzheimer's disease diagnosis using
  evolutionary rvfl classifier.
\newblock {\em IEEE Journal of Biomedical and Health Informatics\/}, (pp.
  1--9).

\bibitem[{Goyal et~al.(2024)Goyal, Rani, \& Singh}]{goyal2023multilayered}
Goyal, P., Rani, R., \& Singh, K. (2024).
\newblock A multilayered framework for diagnosis and classification of
  alzheimer's disease using transfer learned alexnet and lstm.
\newblock {\em Neural Computing and Applications\/}, {\em 36\/}(7), 3777--3801.

\bibitem[{G.S \& S(2023, 10.1109/RMKMATE59243.2023.10369111)}]{GS2023}
G.S, G., \& S, N. (2023, 10.1109/RMKMATE59243.2023.10369111).
\newblock A comprehensive review on early diagnosis of alzheimer’s disease
  detection.
\newblock (pp. 1--6). IEEE.
\newline\urlprefix\url{https://ieeexplore.ieee.org/document/10369111/}

\bibitem[{Han et~al.(2023)Han, Li, Fang, \& Yang}]{10365189}
Han, K., Li, G., Fang, Z., \& Yang, F. (2023).
\newblock Multi-template meta-information regularized network for
  {A}lzheimer’s {D}isease diagnosis using structural {MRI}.
\newblock {\em IEEE Transactions on Medical Imaging\/}, (pp. 1--1).

\bibitem[{Hao et~al.(2024)Hao, Li, Ma, Qin, Zhang, Liu, Initiative
  et~al.}]{hao2023hypergraph}
Hao, X., Li, J., Ma, M., Qin, J., Zhang, D., Liu, F., Initiative, A. D.~N.,
  et~al. (2024).
\newblock Hypergraph convolutional network for longitudinal data analysis in
  {A}lzheimer's {D}isease.
\newblock {\em Computers in Biology and Medicine\/}, {\em 168\/}, 107765.

\bibitem[{Hastie et~al.(2009)Hastie, Tibshirani, Friedman, \&
  Friedman}]{hastie2009elements}
Hastie, T., Tibshirani, R., Friedman, J.~H., \& Friedman, J.~H. (2009).
\newblock {\em The elements of statistical learning: data mining, inference,
  and prediction\/}, vol.~2.
\newblock Springer.

\bibitem[{Hsieh et~al.(2019)Hsieh, Hsiao, Liaw, Huang, \&
  Yang}]{hsieh2019effects}
Hsieh, S.-W., Hsiao, S.-F., Liaw, L.-J., Huang, L.-C., \& Yang, Y.-H. (2019).
\newblock Effects of multiple training modalities in the elderly with
  subjective memory complaints: A pilot study.
\newblock {\em Medicine\/}, {\em 98\/}(29), e16506.

\bibitem[{Huang et~al.(2006)Huang, Zhu, \& Siew}]{huang2006extreme}
Huang, G.-B., Zhu, Q.-Y., \& Siew, C.-K. (2006).
\newblock Extreme learning machine: theory and applications.
\newblock {\em Neurocomputing\/}, {\em 70\/}(1-3), 489--501.

\bibitem[{Huang et~al.(2013)Huang, Shi, \& Suykens}]{6604389}
Huang, X., Shi, L., \& Suykens, J.~A. (2013).
\newblock Support vector machine classifier with pinball loss.
\newblock {\em IEEE Transactions on Pattern Analysis and Machine
  Intelligence\/}, {\em 36\/}(5), 984--997.

\bibitem[{Iosifidis et~al.(2013)Iosifidis, Tefas, \& Pitas}]{6542653}
Iosifidis, A., Tefas, A., \& Pitas, I. (2013).
\newblock Minimum class variance extreme learning machine for human action
  recognition.
\newblock {\em IEEE Transactions on Circuits and Systems for Video
  Technology\/}, {\em 23\/}(11), 1968--1979.

\bibitem[{Iosifidis et~al.(2014)Iosifidis, Tefas, \&
  Pitas}]{iosifidis2014minimum}
Iosifidis, A., Tefas, A., \& Pitas, I. (2014).
\newblock Minimum variance extreme learning machine for human action
  recognition.
\newblock (pp. 5427--5431). \textit{2014 IEEE International Conference on
  Acoustics, Speech and Signal Processing (ICASSP)}.

\bibitem[{Iosifidis et~al.(2016)Iosifidis, Tefas, \& Pitas}]{7052327}
Iosifidis, A., Tefas, A., \& Pitas, I. (2016).
\newblock Graph embedded extreme learning machine.
\newblock {\em IEEE Transactions on Cybernetics\/}, {\em 46\/}(1), 311--324.

\bibitem[{Jacobs et~al.(2018)Jacobs, Hopkins, Mayrhofer, Bruner, van Leeuwen,
  Raaijmakers, \& Schmahmann}]{jacobs2018cerebellum}
Jacobs, H.~I., Hopkins, D.~A., Mayrhofer, H.~C., Bruner, E., van Leeuwen,
  F.~W., Raaijmakers, W., \& Schmahmann, J.~D. (2018).
\newblock The cerebellum in alzheimer’s disease: evaluating its role in
  cognitive decline.
\newblock {\em Brain\/}, {\em 141\/}(1), 37--47.

\bibitem[{Jayadeva et~al.(2007)Jayadeva, Khemchandani, \&
  Chandra}]{khemchandani2007twin}
Jayadeva, Khemchandani, R., \& Chandra, S. (2007).
\newblock Twin support vector machines for pattern classification.
\newblock {\em IEEE Transactions on Pattern Analysis and Machine
  Intelligence\/}, {\em 29\/}(5), 905--910.

\bibitem[{Jessen et~al.(2020)Jessen, Amariglio, Buckley, van~der Flier, Han,
  Molinuevo, Rabin, Rentz, Rodriguez-Gomez, \&
  Saykin}]{jessen2020characterisation}
Jessen, F., Amariglio, R.~E., Buckley, R.~F., van~der Flier, W.~M., Han, Y.,
  Molinuevo, J.~L., Rabin, L., Rentz, D.~M., Rodriguez-Gomez, O., \& Saykin,
  A.~J. (2020).
\newblock The characterisation of subjective cognitive decline.
\newblock {\em The Lancet Neurology\/}, {\em 19\/}(3), 271--278.

\bibitem[{Klein \& Tourville(2012)}]{klein2012101}
Klein, A., \& Tourville, J. (2012).
\newblock 101 labeled brain images and a consistent human cortical labeling
  protocol.
\newblock {\em Frontiers in Neuroscience\/}, {\em 6\/}, 171.

\bibitem[{Kumar \& Gopal(2009)}]{kumar2009least}
Kumar, M.~A., \& Gopal, M. (2009).
\newblock Least squares twin support vector machines for pattern
  classification.
\newblock {\em Expert Systems with Applications\/}, {\em 36\/}(4), 7535--7543.

\bibitem[{Liew(2020)}]{liew2020subjective}
Liew, T.~M. (2020).
\newblock Subjective cognitive decline, anxiety symptoms, and the risk of mild
  cognitive impairment and dementia.
\newblock {\em Alzheimer's Research \& Therapy\/}, {\em 12\/}, 1--9.

\bibitem[{Ma et~al.(2019)Ma, Zhang, Li, \& Tian}]{8723128}
Ma, Y., Zhang, Q., Li, D., \& Tian, Y. (2019).
\newblock Linex support vector machine for large-scale classification.
\newblock {\em IEEE Access\/}, {\em 7\/}, 70319--70331.

\bibitem[{Malik et~al.(2022)Malik, Ganaie, Tanveer, Suganthan, \&
  Initiative}]{9715258}
Malik, A.~K., Ganaie, M.~A., Tanveer, M., Suganthan, P.~N., \& Initiative, A.
  D. N.~I. (2022).
\newblock Alzheimer's disease diagnosis via intuitionistic fuzzy random vector
  functional link network.
\newblock {\em IEEE Transactions on Computational Social Systems\/}, (pp.
  1--12).

\bibitem[{Malik et~al.(2023)Malik, Gao, Ganaie, Tanveer, \&
  Suganthan}]{malik2022random}
Malik, A.~K., Gao, R., Ganaie, M.~A., Tanveer, M., \& Suganthan, P.~N. (2023).
\newblock Random vector functional link network: Recent developments,
  applications, and future directions.
\newblock {\em Applied Soft Computing\/}, {\em 143\/}, 110377.

\bibitem[{Menagadevi et~al.(2024)Menagadevi, Devaraj, Madian, \&
  Thiyagarajan}]{Menagadevi2024}
Menagadevi, M., Devaraj, S., Madian, N., \& Thiyagarajan, D. (2024).
\newblock Machine and deep learning approaches for alzheimer disease detection
  using magnetic resonance images: An updated review.
\newblock {\em Measurement\/}, {\em 226\/}, 114100.
\newline\urlprefix\url{https://linkinghub.elsevier.com/retrieve/pii/S0263224123016640}

\bibitem[{Micheyl et~al.(2007)Micheyl, Carlyon, Gutschalk, Melcher, Oxenham,
  Rauschecker, Tian, \& Wilson}]{micheyl2007role}
Micheyl, C., Carlyon, R.~P., Gutschalk, A., Melcher, J.~R., Oxenham, A.~J.,
  Rauschecker, J.~P., Tian, B., \& Wilson, E.~C. (2007).
\newblock The role of auditory cortex in the formation of auditory streams.
\newblock {\em Hearing Research\/}, {\em 229\/}(1-2), 116--131.

\bibitem[{Miley-Akerstedt et~al.(2019)Miley-Akerstedt, Jelic, Marklund, Walles,
  {\AA}kerstedt, Hagman, \& Andersson}]{miley2019lifestyle}
Miley-Akerstedt, A., Jelic, V., Marklund, K., Walles, H., {\AA}kerstedt, T.,
  Hagman, G., \& Andersson, C. (2019).
\newblock Lifestyle factors are important contributors to subjective memory
  complaints among patients without objective memory impairment or positive
  neurochemical biomarkers for alzheimer’s disease.
\newblock {\em Dementia and Geriatric Cognitive Disorders Extra\/}, {\em
  8\/}(3), 439--452.

\bibitem[{Mitchell(2008)}]{mitchell2008clinical}
Mitchell, A.~J. (2008).
\newblock The clinical significance of subjective memory complaints in the
  diagnosis of mild cognitive impairment and dementia: a meta-analysis.
\newblock {\em International Journal of Geriatric Psychiatry: A Journal of the
  Psychiatry of Late Life and Allied Sciences\/}, {\em 23\/}(11), 1191--1202.

\bibitem[{Molnar(2020)}]{molnar2020interpretable}
Molnar, C. (2020).
\newblock {\em Interpretable machine learning\/}.
\newblock Lulu.com.

\bibitem[{Morrison et~al.(2023)Morrison, Dadar, Villeneuve, Ducharme, \&
  Collins}]{morrison2023white}
Morrison, C., Dadar, M., Villeneuve, S., Ducharme, S., \& Collins, D.~L.
  (2023).
\newblock White matter hyperintensity load varies depending on subjective
  cognitive decline criteria.
\newblock {\em GeroScience\/}, {\em 45\/}(1), 17--28.

\bibitem[{{Muhammed Niyas} \& Thiyagarajan(2023)}]{MuhammedNiyas2023}
{Muhammed Niyas}, K., \& Thiyagarajan, P. (2023).
\newblock A systematic review on early prediction of mild cognitive impairment
  to alzheimer's using machine learning algorithms.
\newblock {\em International Journal of Intelligent Networks\/}, {\em 4\/},
  74--88.
\newline\urlprefix\url{https://www.sciencedirect.com/science/article/pii/S2666603023000052}

\bibitem[{Mujahid et~al.(2023)Mujahid, Rehman, Alam, Alamri, Fati, \&
  Saba}]{mujahid2023efficient}
Mujahid, M., Rehman, A., Alam, T., Alamri, F.~S., Fati, S.~M., \& Saba, T.
  (2023).
\newblock An efficient ensemble approach for {A}lzheimer’s {D}isease
  detection using an adaptive synthetic technique and deep learning.
\newblock {\em Diagnostics\/}, {\em 13\/}(15), 2489.

\bibitem[{Pao et~al.(1994)Pao, Park, \& Sobajic}]{pao1994learning}
Pao, Y.-H., Park, G.-H., \& Sobajic, D.~J. (1994).
\newblock Learning and generalization characteristics of the random vector
  functional-link net.
\newblock {\em Neurocomputing\/}, {\em 6\/}(2), 163--180.

\bibitem[{Potvin et~al.(2019)Potvin, Turcotte, Duchesne
  et~al.}]{potvin2019association}
Potvin, O., Turcotte, V., Duchesne, S., et~al. (2019).
\newblock Association between cerebellum volumes and cognitive functioning.
\newblock {\em Alzheimer Disease \& Associated Disorders\/}.

\bibitem[{Rabin et~al.(2017)Rabin, Smart, \& Amariglio}]{rabin2017subjective}
Rabin, L.~A., Smart, C.~M., \& Amariglio, R.~E. (2017).
\newblock Subjective cognitive decline in preclinical alzheimer's disease.
\newblock {\em Annual Review of Clinical Psychology\/}, {\em 13\/}, 369--396.

\bibitem[{Rao et~al.(2022)Rao, Ganaraja, Murlimanju, Joy, Krishnamurthy, \&
  Agrawal}]{rao2022hippocampus}
Rao, Y.~L., Ganaraja, B., Murlimanju, B., Joy, T., Krishnamurthy, A., \&
  Agrawal, A. (2022).
\newblock Hippocampus and its involvement in alzheimer’s disease: a review.
\newblock {\em 3 Biotech\/}, {\em 12\/}(2), 55.

\bibitem[{Rezvani et~al.(2019)Rezvani, Wang, \& Pourpanah}]{8616852}
Rezvani, S., Wang, X., \& Pourpanah, F. (2019).
\newblock Intuitionistic fuzzy twin support vector machines.
\newblock {\em IEEE Transactions on Fuzzy Systems\/}, {\em 27\/}(11),
  2140--2151.

\bibitem[{Risacher et~al.(2015)Risacher, Kim, Nho, Foroud, Shen, Petersen,
  Jack~Jr, Beckett, Aisen, Koeppe et~al.}]{risacher2015apoe}
Risacher, S.~L., Kim, S., Nho, K., Foroud, T., Shen, L., Petersen, R.~C.,
  Jack~Jr, C.~R., Beckett, L.~A., Aisen, P.~S., Koeppe, R.~A., et~al. (2015).
\newblock Apoe effect on alzheimer's disease biomarkers in older adults with
  significant memory concern.
\newblock {\em Alzheimer's \& Dementia\/}, {\em 11\/}(12), 1417--1429.

\bibitem[{Sajid et~al.(2024)Sajid, Malik, Tanveer, \&
  Suganthan}]{2024sajidnfrvfl}
Sajid, M., Malik, A.~K., Tanveer, M., \& Suganthan, P.~N. (2024).
\newblock Neuro-fuzzy random vector functional link neural network for
  classification and regression problems.
\newblock {\em IEEE Transactions on Fuzzy Systems\/}, (pp. 1--13).

\bibitem[{Saleem et~al.(2022)Saleem, Zahra, Wu, Alwakeel, Alwakeel, Jeribi, \&
  Hijji}]{Saleem2022}
Saleem, T.~J., Zahra, S.~R., Wu, F., Alwakeel, A., Alwakeel, M., Jeribi, F., \&
  Hijji, M. (2022).
\newblock Deep learning-based diagnosis of alzheimer’s disease.
\newblock {\em Journal of Personalized Medicine\/}, {\em 12\/}(5), 815.

\bibitem[{Salman et~al.()Salman, Stephan, \& Hasan}]{Salman}
Salman, A.~W., Stephan, J.~J., \& Hasan, S.~M. (????).
\newblock A comprehensive review on al-zheimer disease.
\newblock {\em Journal of Pharmaceutical Negative Results ¦\/}, {\em 13\/},
  2022.
\newline\urlprefix\url{https://adni.loni.usc.edu/adni/adni/adni/adni/}

\bibitem[{Sanjay \& Swarnalatha(2023)}]{Sanjay2023}
Sanjay, V., \& Swarnalatha, P. (2023).
\newblock An overview of deep learning approaches for alzheimer’s disease
  classification: A review.
\newblock {\em Journal of Advanced Research in Applied Sciences and Engineering
  Technology\/}, {\em 33\/}, 122--140.

\bibitem[{Shaaban et~al.(2023)Shaaban, Antar, Al-Yaman, Mousharafieh, Sabbah,
  Hassan, \& Diab}]{Shaaban2023}
Shaaban, G.~H., Antar, M.~M., Al-Yaman, M.~O., Mousharafieh, N.~I., Sabbah,
  M.~M., Hassan, M., \& Diab, M.~O. (2023).
\newblock From eeg signal to classification in alzheimer disease: A mini
  review.
\newblock In {\em 2023 Seventh International Conference on Advances in
  Biomedical Engineering (ICABME)\/}, (pp. 111--114). IEEE.

\bibitem[{Sharma et~al.(2023{\natexlab{a}})Sharma, Goel, Tanveer, Lin, \&
  Murugan}]{sharma2023deep}
Sharma, R., Goel, T., Tanveer, M., Lin, C.~T., \& Murugan, R.
  (2023{\natexlab{a}}).
\newblock Deep-learning-based diagnosis and prognosis of alzheimer’s disease:
  A comprehensive review.
\newblock {\em IEEE Transactions on Cognitive and Developmental Systems\/},
  {\em 15\/}(3), 1123--1138.

\bibitem[{Sharma et~al.(2022)Sharma, Goel, Tanveer, \& Murugan}]{sharma2022fdn}
Sharma, R., Goel, T., Tanveer, M., \& Murugan, R. (2022).
\newblock F{DN-ADN}et: Fuzzy {LS-TWSVM} based deep learning network for
  prognosis of the {A}lzheimer's {D}isease using the sagittal plane of {MRI}
  scans.
\newblock {\em Applied Soft Computing\/}, {\em 115\/}, 108099.

\bibitem[{Sharma et~al.(2023{\natexlab{b}})Sharma, Goel, Tanveer, Suganthan,
  Razzak, \& Murugan}]{sharma2023conv}
Sharma, R., Goel, T., Tanveer, M., Suganthan, P.~N., Razzak, I., \& Murugan, R.
  (2023{\natexlab{b}}).
\newblock Conv-e{RVFL}: Convolutional neural network based ensemble {RVFL}
  classifier for {A}lzheimer's {D}isease diagnosis.
\newblock {\em IEEE Journal of Biomedical and Health Informatics\/}, {\em
  27\/}(10), 4995--5003.

\bibitem[{Sharma \& Mandal(2022)}]{Sharma2023a}
Sharma, S., \& Mandal, P.~K. (2022).
\newblock A comprehensive report on machine learning-based early detection of
  alzheimer's disease using multi-modal neuroimaging data.
\newblock {\em ACM Computing Surveys (CSUR)\/}, {\em 55\/}(2), 1--44.

\bibitem[{Shi et~al.(2021)Shi, Katuwal, Suganthan, \& Tanveer}]{shi2021random}
Shi, Q., Katuwal, R., Suganthan, P.~N., \& Tanveer, M. (2021).
\newblock Random vector functional link neural network based ensemble deep
  learning.
\newblock {\em Pattern Recognition\/}, {\em 117\/}, 107978.

\bibitem[{Shoeibi et~al.(2023)Shoeibi, Khodatars, Jafari, Ghassemi, Moridian,
  Alizadehsani, Ling, Khosravi, Alinejad-Rokny, Lam, Fuller-Tyszkiewicz,
  Acharya, Anderson, Zhang, \& Gorriz}]{Shoeibi2023}
Shoeibi, A., Khodatars, M., Jafari, M., Ghassemi, N., Moridian, P.,
  Alizadehsani, R., Ling, S.~H., Khosravi, A., Alinejad-Rokny, H., Lam, H.~K.,
  Fuller-Tyszkiewicz, M., Acharya, U.~R., Anderson, D., Zhang, Y., \& Gorriz,
  J.~M. (2023).
\newblock Diagnosis of brain diseases in fusion of neuroimaging modalities
  using deep learning: A review.
\newblock {\em Information Fusion\/}, {\em 93\/}, 85--117.

\bibitem[{Singh et~al.(2022)Singh, D, Soni, \& Kapoor}]{Singh}
Singh, N., D, P., Soni, N., \& Kapoor, A. (2022).
\newblock Automated detection of alzheimer disease using mri images and deep
  neural networks- a review.

\bibitem[{Suykens \& Vandewalle(1999)}]{suykens1999least}
Suykens, J.~A., \& Vandewalle, J. (1999).
\newblock Least squares support vector machine classifiers.
\newblock {\em Neural Processing Letters\/}, {\em 9\/}, 293--300.

\bibitem[{Swords et~al.(2018)Swords, Nguyen, Mudar, \&
  Llano}]{swords2018auditory}
Swords, G.~M., Nguyen, L.~T., Mudar, R.~A., \& Llano, D.~A. (2018).
\newblock Auditory system dysfunction in alzheimer disease and its prodromal
  states: A review.
\newblock {\em Ageing Research Reviews\/}, {\em 44\/}, 49--59.

\bibitem[{Tanveer et~al.(2024)Tanveer, Goel, Sharma, Malik, Beheshti, Del~Ser,
  Suganthan, \& Lin}]{tanveer2024ensemble}
Tanveer, M., Goel, T., Sharma, R., Malik, A.~K., Beheshti, I., Del~Ser, J.,
  Suganthan, P.~N., \& Lin, C. (2024).
\newblock Ensemble deep learning for alzheimer’s disease characterization and
  estimation.
\newblock {\em Nature Mental Health\/}.

\bibitem[{Tanveer et~al.(2020)Tanveer, Richhariya, Khan, Rashid, Khanna,
  Prasad, \& Lin}]{tanveer2020machine}
Tanveer, M., Richhariya, B., Khan, R.~U., Rashid, A.~H., Khanna, P., Prasad,
  M., \& Lin, C. (2020).
\newblock Machine learning techniques for the diagnosis of alzheimer’s
  disease: A review.
\newblock {\em ACM Transactions on Multimedia Computing, Communications, and
  Applications (TOMM)\/}, {\em 16\/}(1s), 1--35.

\bibitem[{Tanveer et~al.(2019)Tanveer, Sharma, \&
  Suganthan}]{tanveer2019general}
Tanveer, M., Sharma, A., \& Suganthan, P.~N. (2019).
\newblock General twin support vector machine with pinball loss function.
\newblock {\em Information Sciences\/}, {\em 494\/}, 311--327.

\bibitem[{Tremblay et~al.(2020)Tremblay, Abbasi, Zeighami, Yau, Dadar, Rahayel,
  \& Dagher}]{tremblay2020sex}
Tremblay, C., Abbasi, N., Zeighami, Y., Yau, Y., Dadar, M., Rahayel, S., \&
  Dagher, A. (2020).
\newblock Sex effects on brain structure in de novo parkinson’s disease: a
  multimodal neuroimaging study.
\newblock {\em Brain\/}, {\em 143\/}(10), 3052--3066.

\bibitem[{Wang et~al.(2024)Wang, Gao, Zhang, \& Han}]{wang2023svfr}
Wang, R., Gao, L., Zhang, X., \& Han, J. (2024).
\newblock Svfr: A novel slice-to-volume feature representation framework using
  deep neural networks and a clustering model for the diagnosis of alzheimer's
  disease.
\newblock {\em Heliyon\/}, {\em 10\/}(1).

\bibitem[{Warren \& Moustafa(2023)}]{Warren2023}
Warren, S.~L., \& Moustafa, A.~A. (2023).
\newblock Functional magnetic resonance imaging, deep learning, and alzheimer's
  disease: A systematic review.
\newblock {\em Journal of Neuroimaging\/}, {\em 33\/}(1), 5--18.

\bibitem[{Warren et~al.(2022)Warren, Reid, Whitfield, \&
  Moustafa}]{warren2022subjective}
Warren, S.~L., Reid, E., Whitfield, P., \& Moustafa, A.~A. (2022).
\newblock Subjective memory complaints as a predictor of mild cognitive
  impairment and alzheimer’s disease.
\newblock {\em Discover Psychology\/}, {\em 2\/}(1), 13.

\bibitem[{Wu et~al.(2021)Wu, Zhang, Li, Kuo, Chen, Dong, Liu, \&
  Yu}]{wu2021cortical}
Wu, B.-S., Zhang, Y.-R., Li, H.-Q., Kuo, K., Chen, S.-D., Dong, Q., Liu, Y., \&
  Yu, J.-T. (2021).
\newblock Cortical structure and the risk for alzheimer’s disease: a
  bidirectional mendelian randomization study.
\newblock {\em Translational Psychiatry\/}, {\em 11\/}(1), 476.

\bibitem[{Wu et~al.(2023)Wu, Hu, Wang, Liu, \& Wang}]{10361545}
Wu, L., Hu, S., Wang, D., Liu, C., \& Wang, L. (2023).
\newblock R{C}la{N}et: An explainable {A}lzheimer's {D}isease diagnosis
  framework by joint registration and classification.
\newblock {\em IEEE Journal of Biomedical and Health Informatics\/}, (pp.
  1--12).

\bibitem[{Yates et~al.(2017)Yates, Clare, Woods, \& CFAS}]{yates2017subjective}
Yates, J.~A., Clare, L., Woods, R.~T., \& CFAS, M. (2017).
\newblock Subjective memory complaints, mood and mci: a follow-up study.
\newblock {\em Aging \& mental health\/}, {\em 21\/}(3), 313--321.

\bibitem[{Zhang \& Suganthan(2016)}]{zhang2016comprehensive}
Zhang, L., \& Suganthan, P.~N. (2016).
\newblock A comprehensive evaluation of random vector functional link networks.
\newblock {\em Information Sciences\/}, {\em 367\/}, 1094--1105.

\end{thebibliography}
\bibliographystyle{apa}

\end{document}